\documentclass{article}

\PassOptionsToPackage{numbers,compress}{natbib}

\usepackage[preprint]{neurips_2022}

\usepackage[utf8]{inputenc}
\usepackage[T1]{fontenc}
\usepackage{hyperref}       
\usepackage{url}            
\usepackage{booktabs}
\usepackage{amsfonts}       
\usepackage{nicefrac}       
\usepackage{microtype}      
\usepackage{xcolor}

\usepackage{enumerate}
\usepackage{bm}
\usepackage{amsmath}
\usepackage{amssymb}
\usepackage{graphicx}
\usepackage{float}
\usepackage{subcaption}
\usepackage{amsthm}
\theoremstyle{definition}

\usepackage[ruled,vlined]{algorithm2e}
\usepackage{cleveref}
\usepackage{mathtools}
\usepackage{multirow}

\usepackage{titlesec}
\titlespacing\section{0pt}{6pt plus 2pt minus 2pt}{0pt plus 2pt minus 2pt}
\titlespacing\subsection{0pt}{3pt plus 1pt minus 1pt}{0pt plus 1pt minus 1pt}

\newtheorem{assumption}{Assumption}
\newtheorem{theorem}{Theorem}
\newtheorem{proposition}[theorem]{Proposition}
\newtheorem{corollary}[theorem]{Corollary}
\theoremstyle{remark}
\newtheorem*{remark}{Remark}

\def\bfu{\bm{u}}
\def\bfv{\bm{v}}

\def\bfx{\bm{x}}
\def\bfy{\bm{y}}

\def\bfT{\bm{T}}

\def\bfbeta{\bm{\beta}}
\def\bfgamma{\bm{\gamma}}
\def\bfdelta{\bm{\delta}}
\def\bfepsilon{\bm{\epsilon}}

\def\bftheta{\bm{\theta}}
\def\bfmu{\bm{\mu}}

\def\bfxi{\bm{\xi}}

\def\bfrho{\bm{\rho}}
\def\bfsigma{\bm{\sigma}}
\def\bftau{\bm{\tau}}

\def\bfGamma{\bm{\Gamma}}
\def\bfTheta{\bm{\Theta}}

\def\bfSigma{\bm{\Sigma}}

\def\bbI{\mathbb I}
\def\bbN{\mathbb N}
\def\bbR{\mathbb R}

\def\E{\operatorname{E}}
\def\Var{\operatorname{Var}}

\def\diag{\operatorname{diag}}

\def\iidsim{\overset{\text{iid}}{\sim}}

\def\Normal{\mathcal{N}}

\def\InvGamma{\mathcal{IG}}
\def\MVN{\mathcal{MVN}}
\def\TN{\mathcal{TN}}

\def\rest{\mathrm{rest}}
\def\test{\mathrm{test}}
\def\train{\mathrm{train}}

\def\inf{$\infty$}
\def\mp{$\pm$}

\title{
  Density Regression and Uncertainty Quantification with
  Bayesian Deep Noise Neural Networks
}

\author{%
  Daiwei Zhang \\
  Department of Biostatistics \\
  University of Pennsylvania \\
  Philadelphia, PA 19104 \\
  \texttt{daiwei.zhang@pennmedicine.upenn.edu} \\
  \And
  Tianci Liu \\
  School of Electrical and Computer Engineering \\
  Purdue University \\
  West Lafayette, IN 47907 \\
  \texttt{lliutianc@purdue.edu} \\
  \And
  Jian Kang\thanks{Corresponding author} \\
  Department of Biostatistics \\
  University of Michigan \\
  Ann Arbor, MI 48105 \\
  \texttt{jiankang@umich.edu}
}

\begin{document}

\maketitle

\begin{abstract}

Deep neural network (DNN) models have achieved state-of-the-art predictive accuracy
in a wide range of supervised learning applications.
However, accurately quantifying the uncertainty in DNN predictions
remains a challenging task.
For continuous outcome variables,
an even more difficult problem is to 
estimate the predictive density function,
which not only provides a natural quantification
of the predictive uncertainty,
but also fully captures the random variation in the outcome.
In this work, we propose the
Bayesian Deep Noise Neural Network (B-DeepNoise),
which generalizes standard Bayesian DNNs
by extending the random noise variable
from the output layer to all hidden layers.
The latent random noise
equips B-DeepNoise with the flexibility to approximate
highly complex predictive distributions
and accurately quantify predictive uncertainty.
For posterior computation,
the unique structure of B-DeepNoise leads to
a closed-form  Gibbs sampling algorithm
that iteratively simulates from
the posterior full conditional distributions
of the model parameters,
circumventing computationally intensive
Metropolis-Hastings methods.
A theoretical analysis of B-DeepNoise
establishes a recursive representation
of the predictive distribution
and decomposes the predictive variance
with respect to the latent parameters.
We evaluate B-DeepNoise against existing methods
on benchmark regression datasets,
demonstrating its superior performance
in terms of prediction accuracy,
uncertainty quantification accuracy,
and uncertainty quantification efficiency.
To illustrate our method's usefulness
in scientific studies,
we apply B-DeepNoise to
predict general intelligence from neuroimaging features
in the Adolescent Brain Cognitive Development (ABCD) project.

\end{abstract}

\section{Introduction}

Deep neural networks (DNNs)
\citep{goodfellow2016deep,lecun2015deep}
have achieved outstanding prediction performance
in a wide range of artificial intelligence (AI) applications
\citep{
    pouyanfar2018survey,berner2021modern,bojarski2016end,%
    grigorescu2020survey,ker2017deep,zou2019primer,%
    stephenson2019survey,zemouri2019deep}.
Despite overwhelming cases of success,
a major drawback of standard DNNs is
the lack of reliable
uncertainty quantification (UQ) \citep{begoli2019need}.
UQ is an essential task in safety-critical AI applications
\citep{amodei2016concrete}.
For example, in medical diagnosis,
an individualized risk assessment AI model
should be able to report its confidence in its predictions.
When the AI model is not sufficiently certain in its assessment of a patient,
the patient should be referred to human physicians for further evaluation
\citep{jiang2012calibrating,leibig2017leveraging}.

In this work, we seek to solve the problem of UQ in DNN regression tasks.
In a standard DNN regression model,
the outcome $y_i$ and the predictors $\bfx_i$ 
are assumed to follow the relation
$y_i = f(\bfx_i) + \epsilon_i$, 
where the mean function $f(\bfx_i) = \mathrm{E}(y_i \mid \bfx_i)$
is constructed by a DNN,
and the random noise $\epsilon_i$
follows a zero-mean homoscedastic Gaussian distribution
$\epsilon_i \iidsim \Normal(0, \sigma^2)$
for some unknown $\sigma^2>0$.
This formulation implies the conditional variance
of the outcome variable $\Var(y_i\mid \bfx_i)$,
given the predictor, to be constant.
However, in real applications, the true conditional distribution
could be heteroscedastic Gaussian
(i.e. $\Var(y_i\mid \bfx_i) = \sigma^2(\bfx_i)$)  
or not Gaussian at all
(e.g. the distribution of $\epsilon_i$ is asymmetric or multimodal).
In these cases, common UQ statistics derived from Gaussian models
(e.g. predictive variance)
may fail to capture important patterns in the data,
causing UQ to be misleading or inefficient.
Therefore, in order to achieve accurate UQ in DNN regression problems,
it is critical to learn the conditional density
of the outcome variable given the predictors,
since the conditional density
not only fully quantifies prediction uncertainty,
but also gives a complete picture of possible variation
in the outcome.
We refer to the problem of estimating the conditional density function
given the predictors
as the \emph{density regression} (DR) problem,
following the terminology in \citep{dunson2007bayesian}.
In this work, we focus on DR tasks with DNNs
and treat UQ as a byproduct of DR.

\subsection{Related Work}
To estimate the conditional density of the outcome given the predictors,
many DNN-based frequentist DR methods have been proposed
\citep{abdar2021review,staahl2020evaluation,caldeira2020deeply,zhu2019physics}.
A conceptually straightforward DR method is to
extend the conditional distribution of the outcome
from Gaussian to Gaussian mixture,
such as in mixture density networks
\citep{bishop1994mixture,bishop2006pattern}
and deep ensembles \citep{lakshminarayanan2016simple}.
An alternative approach is to estimate the prediction intervals
without distribution assumptions on the outcome
\citep{lei2014distribution,pearce2018high},
which is closely related to quantile-based models
\citep{romano2019conformalized,tagasovska2019single}.
Other solutions include converting the continuous outcome
into a multi-class categorical variable by using bins
\citep{li2021deep}
and estimating the cumulative distribution function (CDF) directly
\citep{huberman2021nonparametric}.

Compared to ad hoc frequentist UQ and DR methods,
the Bayesian framework for DNNs,
or Bayesian neural networks (BNNs) \citep{mackay1995probable,neal2012bayesian,xue2019reliable},
provides a more natural and systematic way to model uncertainty in the outcome,
i.e. by using the posterior predictive distribution.
Theoretically, the posterior prediction intervals
are well-calibrated asymptotically
\citep{hwang1997prediction,wang2020uncertainty,sun2021sparse}.
In addition to the capacity of UQ, the Bayesian framework
improves the prediction accuracy of deterministic DNNs
\citep{kendall2017uncertainties,izmailov2018averaging}.
Regardless of the aptness of BNNs for UQ,
due to the intractability of their posterior distributions,
one must resort to variational inference (VI)
or Markov Chain Monte Carlo (MCMC) simulation
for posterior computation.
VI methods \citep{blei2017variational,kingma2013auto,mandt2017stochastic}
approximate the posterior distribution with simpler distributions
\citep{
    graves2011practical,louizos2016structured,lee2020estimating,%
    rezende2015variational,louizos2017multiplicative}.
Common randomness-based regularization techniques for deterministic DNNs,
such as dropout
\citep{srivastava2014dropout,gal2016dropout,molchanov2017variational},
batch normalization
\citep{ioffe2015batch,teye2018bayesian},
and random weights
\citep{blundell2015weight,hernandez2015probabilistic},
can be interpreted as special cases of VI.
However, although computationally efficient,
VI methods induce extra approximation errors
in learning the posterior distribution,
with the parameter variance and covariance
often underestimated or oversimplified
\citep{blei2017variational}.

In contrast, MCMC methods simulate
the \emph{exact} posterior distribution of the BNN.
The most popular MCMC algorithm for modern applications
is arguably the Metropolis-Hastings (MH) algorithm
\citep{chib1995understanding,hitchcock2003history,andrieu2008tutorial}.
However, even with the assistance
of efficient techniques such as
Hamiltonian dynamics \citep{wenzel2020good,wilson2020bayesian},
Langevin dynamics \citep{welling2011bayesian},
stochastic gradients \citep{chen2014stochastic,chen2016bridging},
and mini-batches \citep{wu2020mini},
the ultrahigh dimensionality of BNNs
has caused prohibitive computation cost for MH algorithms
\citep{liang2016bootstrap,jospin2022hands}.
An alternative MCMC simulation method is Gibbs sampling
\citep{geman1984stochastic,gelfand1990sampling,roberts1994simple},
where each model parameter (or a block of parameters)
is sampled from the conditional posterior distribution
given all the other parameters.
Although block-wise Gibbs sampling solves
the so-called curse of dimensionality
for models with a high number of parameters
\citep{gelfand2000gibbs},
its applications to standard BNNs are impractical if not impossible,
due to the lack of closed-form full conditional posterior distributions
in standard BNNs with nonlinear activation functions.
Finally, most existing BNN methods (including both MCMC- and VI-based)
focus on UQ,
whereas the problem of DR with DNN is scarcely studied in the Bayesian framework
\citep{dunson2007bayesian}.

For UQ and DR with DNN,
a related topic is the incorporation of latent noise in the hidden layers.
However, existing works on latent noise in DNNs primarily use it
as a means for regularization
\citep{you2019adversarial,gulcehre2016noisy}.
The potential of stochastic activation layers for UQ
was briefly discussed in \citep{lee2019probact},
but the context of this work was classification tasks,
where the predictive uncertainty could already be fully characterized by
well-calibrated categorical distributions without using any latent noise.
More recently, \citep{sun2022kernel}
formulated DNNs as latent variable models and
included kernel maps in the input layer to avoid feature collinearity.
Although the proposed model was capable of UQ,
the more challenging problem of DR was not studied.


\subsection{Our Contributions}
To address the existing issues associated with DR and UQ for DNNs,
we propose the Bayesian Deep Noise Neural Network (B-DeepNoise).
B-DeepNoise generalizes standard BNNs
by adding latent random noise
both before and after every activation layer.
Although the latent random noise variables
independently follow Gaussian distributions,
their composition across multiple
layers with non-linear activations
can generate highly complex predictive density functions.
Moreover, the unique structure of B-DeepNoise
induces closed-form full conditional posterior distributions
for the model parameters,
which eliminates the primary barrier for Gibbs sampling in DNN-based models
and therefore makes it possible
to simulate the exact posterior distribution
without using computationally intensive MH algorithms.

To our best knowledge, this is the first work on
estimating complex predictive density functions
by utilizing DNNs with latent random noise.
Furthermore, no previous work has developed Gibbs sampling algorithms
for DNN-based Bayesian models.
In short, our work contributes to the existing literature
on DR and UQ with DNNs
in the following ways: 
\begin{itemize}
\item We propose a Bayesian DNN model that is capable of learning
a wide range of complex (e.g. heteroscedastic, asymmetric, multimodal)
predictive density functions.
\item We develop a Gibbs sampling algorithm for
the posterior computation of our model
that can be implemented by using common samplers and
without the need of MH steps.
\item We perform theoretical analysis of our method and obtain
analytic expressions of the predictive density and variance propagation.
\item We compare our model with existing methods on synthetic and real benchmark datasets
and demonstrate its usefulness in a neuroimaging study.
\end{itemize}


\section{Model Description}
\label{sec:method}


\subsection{DNNs with latent noise variables}
\label{sec:model}

\begin{figure}[t]
  \scriptsize
  \captionsetup{font=footnotesize}
  \caption{
    Observed data (blue dots)
    and estimated predictive density (heatmap)
    by B-DeepNoise (top row), DE (middle row), and VI (bottom row)
    for heteroscedastic (left column), asymmetric (middle column),
    and multimodal (right column) noise.
    \label{fig:simu}
  }
  \includegraphics[height=0.9in]{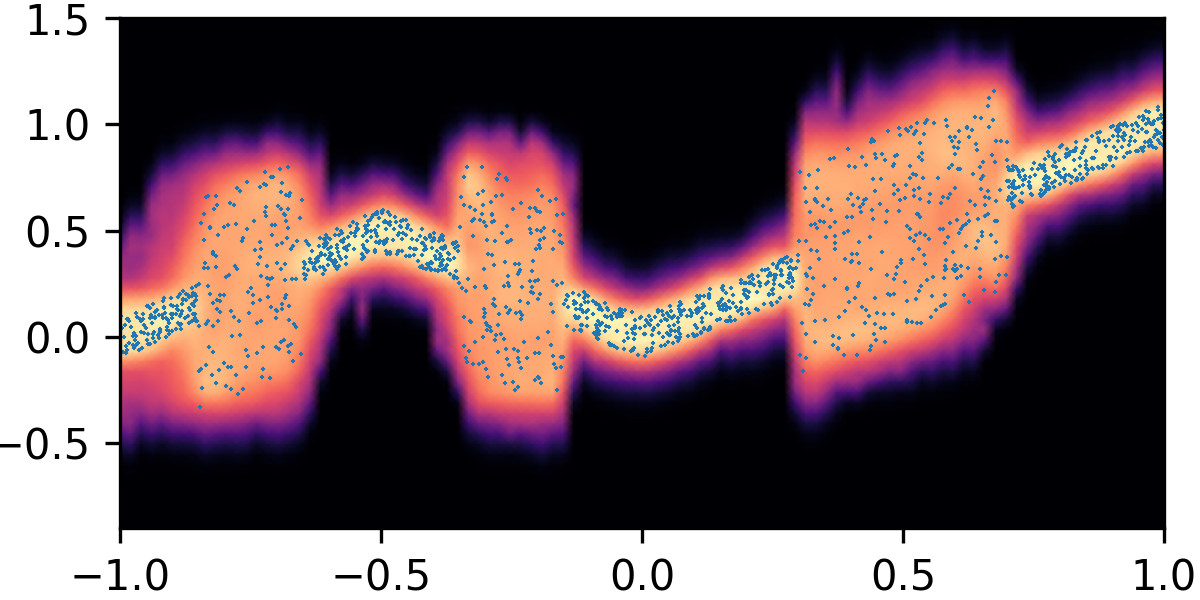}
  \includegraphics[height=0.9in]{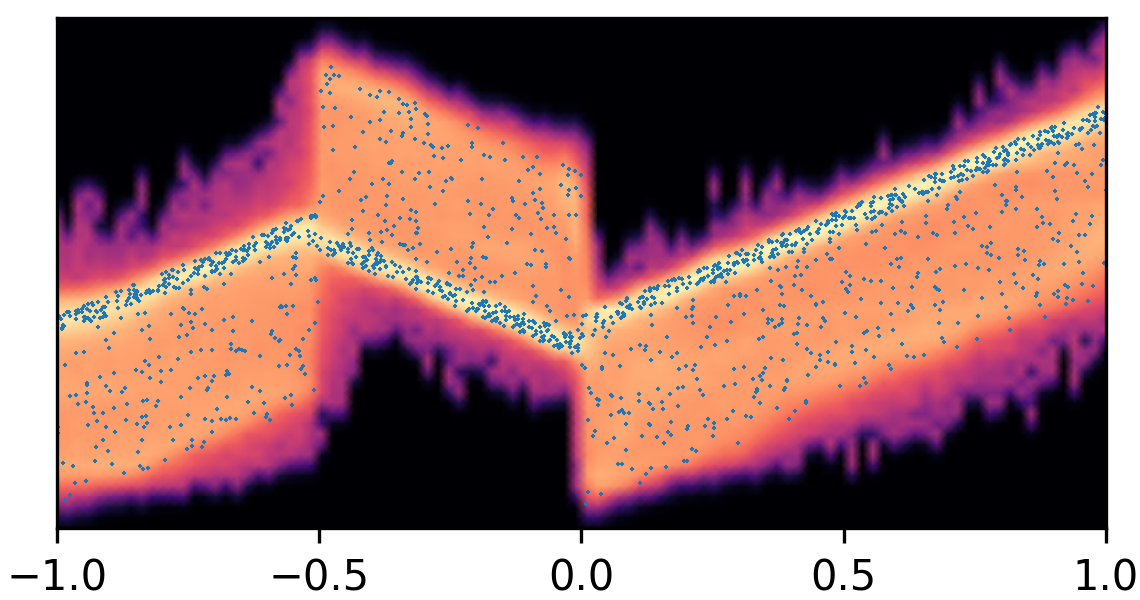}
  \includegraphics[height=0.9in]{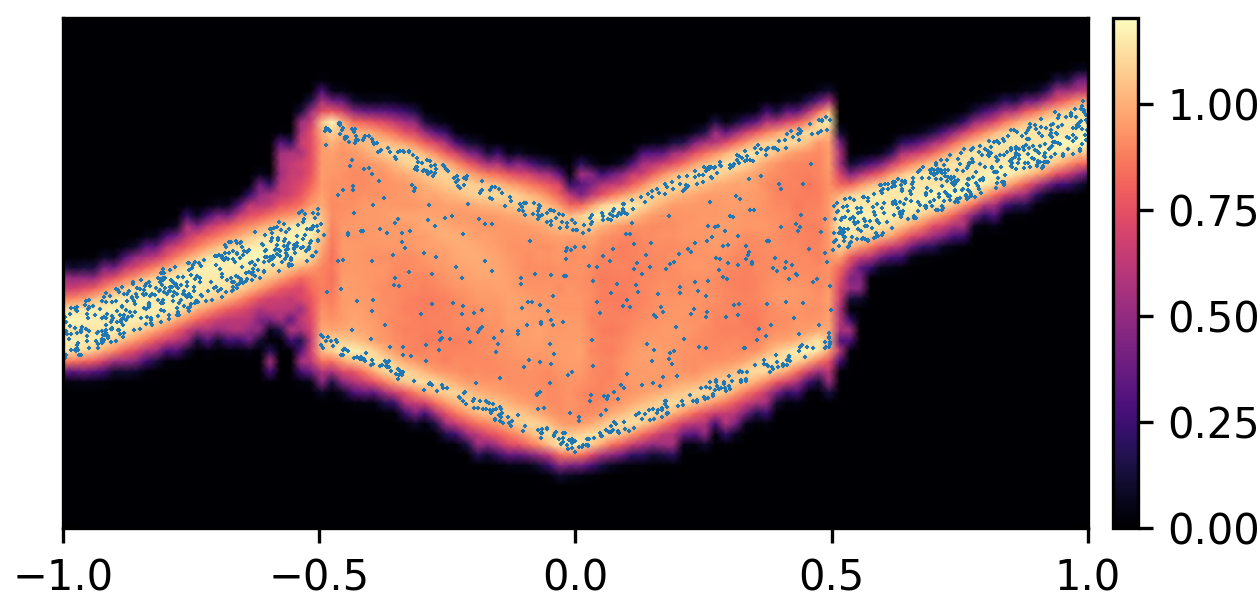}
  \\
  \includegraphics[height=0.9in]{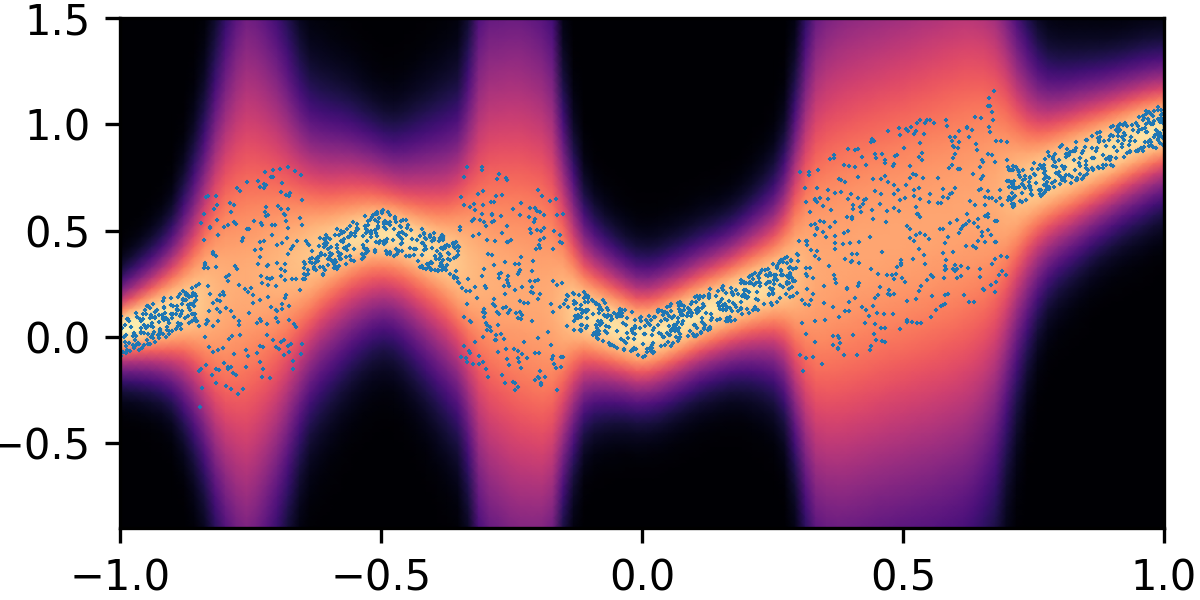}
  \includegraphics[height=0.9in]{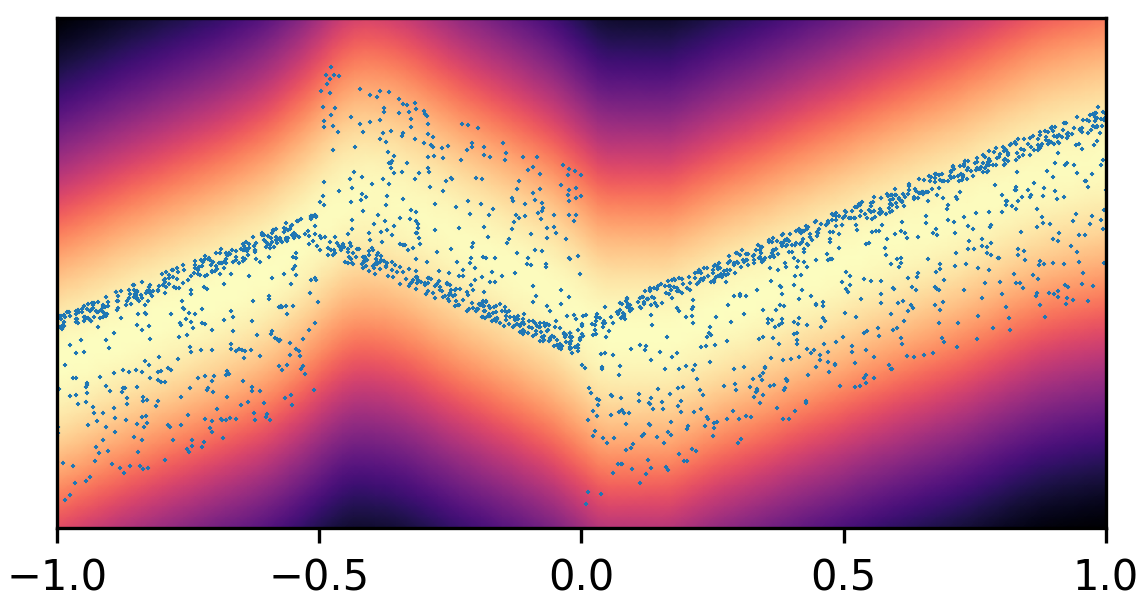}
  \includegraphics[height=0.9in]{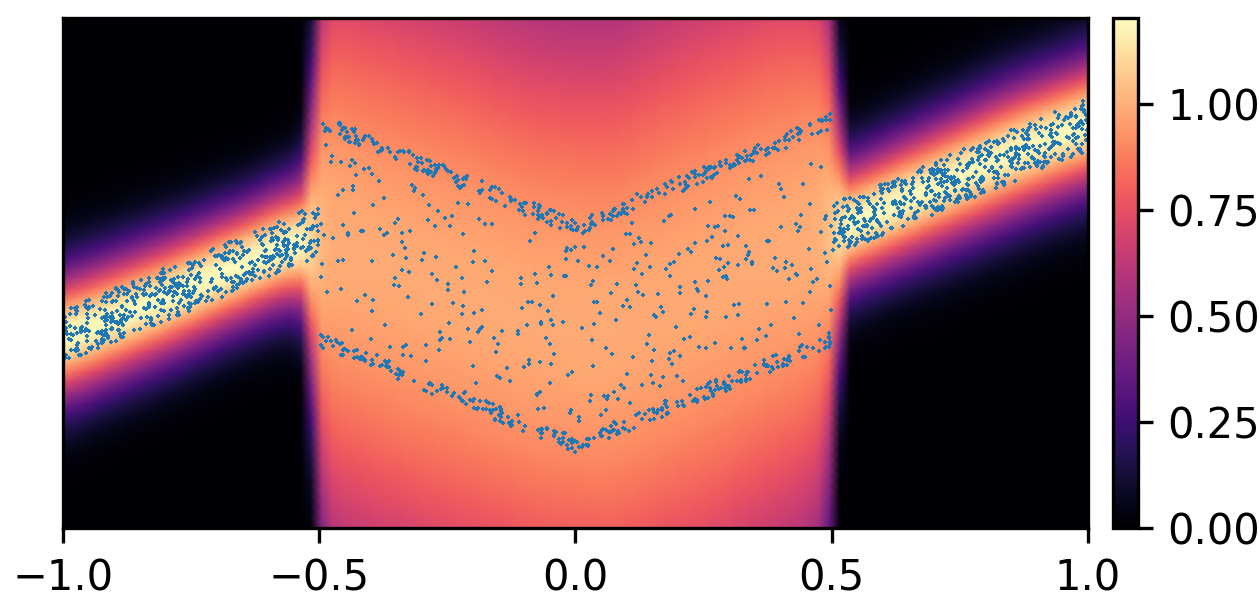}
  \\
  \includegraphics[height=0.9in]{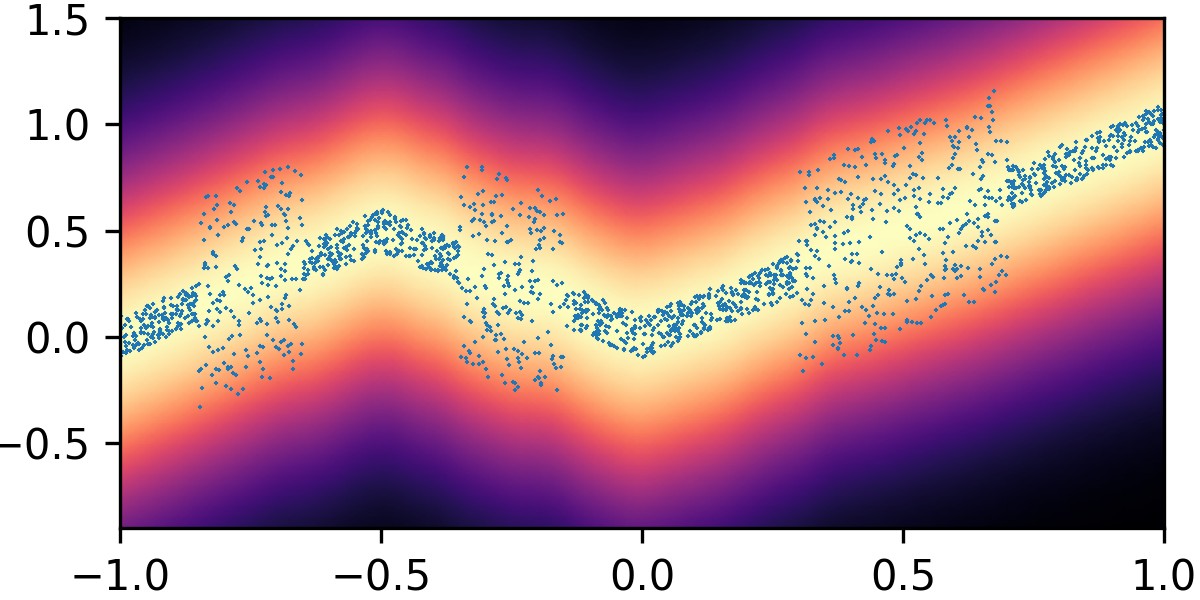}
  \includegraphics[height=0.9in]{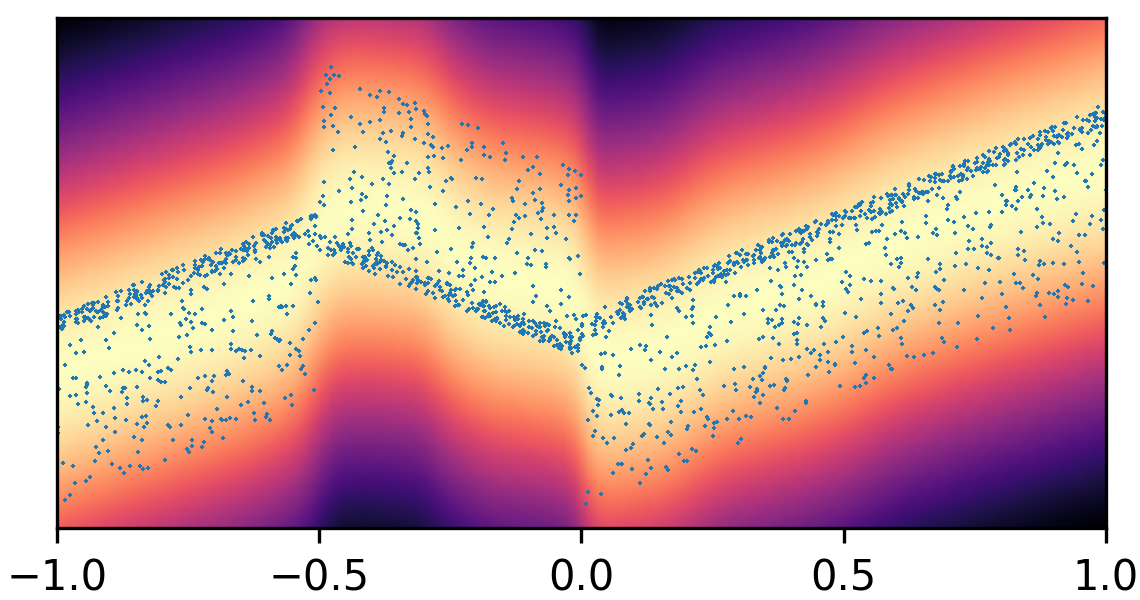}
  \includegraphics[height=0.9in]{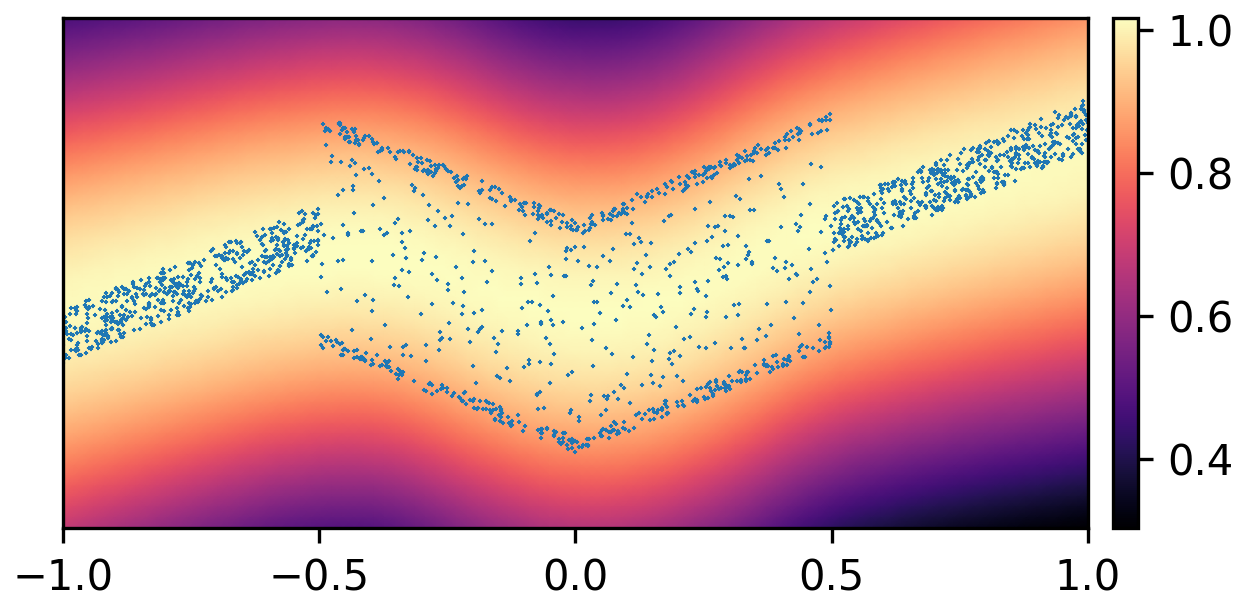}
\end{figure}

Suppose the data consist of $N$ observations. For $n =1,\ldots, N$, let $\bfx^{(n)} \in \bbR^{P}$ be the predictors  and  $\bfy^{(n)} \in \bbR^{Q}$ be the outcome variable.
Let $\Normal(\bfmu,\bfSigma)$ be a Gaussian distribution with mean $\bfmu$ and covariance $\bfSigma$.
To specify the nonlinear association between 
$\bfy^{(n)}$ and $\bfx^{(n)}$, a standard $L$-layer feed-forward DNN model with Gaussian noises can be represented as
\begin{align}
  \label{eq:dnn-endpoints}
  &[\bfy^{(n)}\mid \bfu_L^{(n)}]
  \iidsim 
  \Normal\left(\bfbeta_L\bfu^{(n)}_{L} + \bfgamma_L, \bfT_L\right),\qquad 
   \bfu^{(n)}_0
  = 
  \bfx^{(n)},\\
   &\bfu^{(n)}_{l+1}
  = 
  h\Big(\bfbeta_{l} \bfu^{(n)}_{l} + \bfgamma_{l} \Big), \qquad  l \in \{1,\ldots, L-1\}
  \label{eq:dnn-hidden}
\end{align}
where $\bfbeta_l \in \bbR^{{K_l} \times K_{l-1}}$ and
$\bfgamma_l \in \bbR^{K_l}$ are unknown parameters,
$K_l$ is the number of units in the $l$th layer,
and $h(\cdot)$ is an element-wise nonlinear activation function,
such as the rectified linear unit (ReLU) function and logistic function.
In this formulation, $\bfu^{(n)}_L$ is a deterministic function of $\bfx^{(n)}$. This implies that $\bfy^{(n)}$ given $\bfx^{(n)}$ follows the same conditional distribution as $\bfy^{(n)}$ given $\bfu_L^{(n)}$, i.e., a homoscedastic Gaussian distribution with constant covariance $\bfT_L$. 
To model more complex conditional distribution,
we propose the Deep Noise Neural Network (DeepNoise),
which generalizes \Cref{eq:dnn-hidden} of the standard DNN model by
including noise variables before and after every activation layer:
for $l \in \{1,\ldots, L-1\}$,  
\begin{align}
  \bfu^{(n)}_{l+1}
  = 
  h\Big(\bfbeta_{l} \bfu^{(n)}_{l} + \bfgamma_{l} + \bfepsilon^{(n)}_l \Big) + \bfdelta^{(n)}_l,\qquad \bfepsilon^{(n)}_l
  \iidsim 
  \Normal\Big(\bm{0},\; \bfT_l\Big)
  , \qquad 
  \bfdelta^{(n)}_l
  \iidsim
  \Normal\Big(\bm{0},\ \bfSigma_l\Big),
   \label{eq:dalea-hidden}
\end{align}
where $\bfT_l = \diag\{\bftau^2_l\}$ and  $\bfSigma_l=\diag\{\bfsigma^2_l\}$ with $\bftau^2 = (\tau^2_{l,1},\ldots,\tau^2_{l,K_l})^\top$ and $\bfsigma^2_l = (\sigma^2_{l,1},\ldots, \sigma^2_{l,K_l})^\top$.  By composing the latent Gaussian noise variables
with linear maps and non-linear activations,
DeepNoise is capable of representing
a wide range of heteroscedastic Gaussian and non-Gaussian conditional density functions
(e.g. asymmetric, multimodal),
as illustrated in \Cref{fig:simu}.
Intuitively, as Gaussian mixtures are universal approximators
of densities
(\citep{plataniotis2017gaussian},
\citep{calcaterra2008approximating},
\citep[Sec. 3.9.6]{goodfellow2016deep})
and DNNs are universal approximators of functions
\citep{scarselli1998universal,yarotsky2017error,lu2020universal},
DeepNoise is designed to be a universal approximator of conditional densities.
The nonparametric nature of DeepNoise enables it to approximate
increasingly complex conditional density functions
by increasing the number of hidden layers
and the number of nodes per layer.

\subsection{Model Representation and Prior Specifications}
DeepNoise transforms the predictor vector into the outcome vector
by iteratively applying the linear-noise-nonlinear-noise maps.
Thus, the DeepNoise model defined by combining
\Cref{eq:dnn-endpoints,eq:dalea-hidden} is equivalent to:
\begin{align}
    \label{eq:v}
    \Big[ \bfv_{l}^{(n)} \mid \bfu^{(n)}_l \Big]
    \; \iidsim \;
    \Normal & \bigg(
        \bfbeta_{l} \bfu^{(n)}_l
        +
        \bfgamma_{l}
        , \;
        \bfT_l
    \bigg)
    , &&
    \text{for }
    l = 0, \ldots, L
    \\
    \label{eq:u}
   \Big[ \bfu^{(n)}_{l+1} \mid  \bfv_{l}^{(n)} \Big]
    \; \iidsim \;
    \Normal & \bigg(
        h\Big(\bfv^{(n)}_{l}\Big)
        , \;
        \bfSigma_{l}
    \bigg), &&
    \text{for }
    l = 0, \ldots, L-1
\end{align}
where
$\bfu^{(n)}_0 = \bfx^{(n)}$
and
$\bfv^{(n)}_L = \bfy^{(n)}$. Write $\bfbeta_l=(\beta_{l,k,k'})$ and $\bfgamma_l=(\gamma_{l,k})$. 
To make fully Bayesian inference, we impose normal-inverse-gamma prior distributions
on the weight-bias parameters:
\begin{align*}
  \beta_{l,k,k'} \iidsim \Normal(0, \rho_{l,k,k'}^2)
  , \quad 
  \gamma_{l,k} \iidsim \Normal(0, \xi_{l,k}^2)
  , \quad
  \rho_{l,k,k'}^2 \iidsim \InvGamma(a_{\text{weight}}, b_{\text{weight}})
  , \quad
  \xi_{l,k}^2 \iidsim \InvGamma(a_{\text{bias}}, b_{\text{bias}})
\end{align*}
and assign inverse-gamma prior distributions
to the pre-activation and post-activation noise variances:
\begin{align*}
  \tau_{l,k}^2 \iidsim \InvGamma(a_{\text{preact}}, b_{\text{preact}}),
  \qquad
  \sigma_{l,k}^2 \iidsim \InvGamma(a_{\text{postact}}, b_{\text{postact}}).
\end{align*}
We refer to \eqref{eq:v} and \eqref{eq:u}
along with the prior specifications above as
the Bayesian Deep Noise Neural Network (B-DeepNoise) model.
Although DeepNoise
(i.e. \Cref{eq:v,eq:u} without the prior distributions)
is an effective frequentist model,
for the rest of the paper, we focus on B-DeepNoise
and conduct theoretical, computational, and empirical analyses
in the Bayesian framework.
\subsection{Posterior Computation}
\label{sec:sampling}

Compared to standard DNNs and BNNs,
the addition of latent random noises not only
provides B-DeepNoise with more flexibility
for approximating complex predictive density functions,
but also provides closed-form expressions
of the full conditional posterior distributions
of the model parameters.
The latter advantage makes it possible
to derive efficient Gibbs sampling algorithms for B-DeepNoise.
The model parameters are naturally grouped into five categories:
pre-activation random noises $\bfv^{(n)}_l$,
post-activation random noises $\bfu^{(n)}_l$,
weights and biases $\bfbeta_l, \bfgamma_l$,
weight and bias variances $\bfrho^2_l, \bfxi^2_l$,
and random noise variances $\bftau^2_l, \bfsigma^2_l$.
We derive the full conditional posterior distributions
for each of these groups of parameters.

The most complex parameter group is
the pre-activation random noises $\bfv^{(n)}_l$,
since their full conditional posterior distributions
involves the nonlinear activation function.
To develop straightforward Gibbs samplers,
we require the activation function
to be piecewise linear, as defined in \Cref{assu:activation}.
\begin{assumption}
  \label{assu:activation}
  The activation function can be expressed as
  $
      h(t) = \sum_{j=1}^J
      (b_j t + b'_j)
      \cdot
      \bbI\{t \in [c_{j-1}, c_j) \}
  $
  for some $J \in \bbN_+$,
  $b_{1}, \ldots, b_{J}, b'_{1}, \ldots, b'_{J} \in \bbR$,
  and
  $-\infty = c_0 < c_1 < \ldots < c_{J-1} < c_{J} = \infty$.
\end{assumption}

\begin{remark}
  The family of functions defined in \Cref{assu:activation}
  includes many common activation functions,
  such as ReLU and leaky ReLU \citep{maas2013rectifier}.
  Moreover, smooth activation functions can be approximated
  by piecewise linear functions.
  For example, the logistic, tanh, and softplus functions
  can be approximated by the hard sigmoid, hard tanh,
  and ReLU functions, respectively.
\end{remark}

We now derive the full conditional posterior distributions
of the pre-activation latent noise. 
Let $\TN(\lambda, \omega^2, c_a, c_b)$ be a truncated normal distribution
on interval $[c_a,c_b]$ with location $\lambda$ and scale $\omega$.
Let $\phi_{\Normal}(\cdot \mid \mu, \sigma^2)$ be the PDF of the normal distribution with mean $\mu$ and variance $\sigma^2$. Let $\Phi_{\Normal}(\cdot)$ be the CDF of the standard normal distribution.

\begin{theorem}
  \label{thm:distribution-v}
  Suppose the activation function $h$ satisfies \Cref{assu:activation}.
  For $n \in \{1, \ldots, N\}$, $l\in\{0,\ldots,L\}$, and $k\in\{1,\ldots, K_l\}$,
  let
  $
    \omega_{n,l,k,j}^{2} = \Big(\tau_{l,k}^{-2} + \sigma_{l,k}^{-2} b_j^2\Big)^{-1}
  $
  and \\
  $
    \lambda_{n,l,k,j}
    =
    \tau_{l,k}^{-2}\omega_{n,l,k,j}^{2} \Big(\gamma_{l,k} + \bfbeta_l \bfu_l^{(n)}\Big)
    +
    \sigma_{l,k}^{-2} b_j\omega_{n,l,k,j}^{2} \Big(u^{(n)}_{l+1,k} - b'_j\Big)
  $. \\
  Define $\zeta_{n,l,k,1} = 1$, $\zeta_{n,l,k,j}=\zeta_{n,l,k,j-1} \zeta'_{n,l,k,j}$,
  where
  $
    \zeta'_{n,l,k,j} = \kappa_{j, j-1} \kappa_{j-1,j-1}^{-1} \tilde{\kappa}_{j-1, j-1} \tilde{\kappa}_{j, j-1}^{-1}
  $
  with
  $
    \kappa_{j, j'} = \phi_{\Normal}(u^{(n)}_{l+1,k} \mid c_{j'} b_{j} + b'_{j}, \sigma_{l,k}^2)
  $
  and
  $
    \tilde{\kappa}_{j, j'} = \phi_{\Normal}(c_{j'} \mid \lambda_{n,l,k,j}, \omega_{n,l,k,j}^2)
  $.
  Moreover, let
  $
    c'_{j,j'} = (c_{j'} - \lambda_{n,l,k,j}) \omega_{n,l,k,j}^{-1}
  $,
  $
    \pi'_{n,l,k,j}
    = 
    \Phi_{\Normal}(c'_{j,j})
    -
    \Phi_{\Normal}(c'_{j,j-1})
  $
  and $\pi_{n,l,k,j} \propto \pi'_{n,l,k,j} \zeta_{n,l,k,j}$.
  The conditional posterior distribution
  of pre-activation noise $v^{(n)}_{l,k}$
  given the rest of the parameters is
  \begin{align}
    \label{eq:distribution-v}
    \Big[ v_{l,k}^{(n)} \Bigm| \rest \Big]  \sim & \sum_{j=1}^J \pi_{n,l,k,j}
    \cdot
    \TN\Big(\lambda_{n,l,k,j},\;
      \omega_{n,l,k,j}^2,\;
      c_{j-1}, \; c_j
    \Big).
  \end{align}
\end{theorem}

\begin{remark}
  Intuitively, the piecewise linear property
  of the activation function
  causes the full conditional posterior distribution
  of $\bfv^{(n)}_l$ to be ``piecewise normal'',
  i.e. a mixture of truncated normal distributions
  with adjacent truncation endpoints.
  By using \Cref{thm:distribution-v},
  $\bfv^{(n)}_l$
  can be easily simulated by using samplers
  for categorical distributions
  and truncated normal distributions,
  which are widely available in scientific computation software.
  In addition, the number of mixing components,
  which depends on the activation function,
  is usually very small
  (e.g. 2 for ReLU and 3 for hard tanh).
\end{remark}

The full conditional posterior distributions
of the rest of the model parameters
are either normal or inverse-gamma,
because of the conjugate priors,
as shown in \Cref{thm:distribution-others} in~\Cref{sec:gibbs}.
The derivations are similar to
those for Bayesian linear regression with conjugate priors
\citep[Sec. 2.3.3]{bishop2006pattern}. The complete algorithm is described in \Cref{alg:sampling} in~\Cref{sec:gibbs}.
Note that the predictive distribution only depends on
the weight-bias parameters
and the latent noise variance parameters.
To speed up the computation,
these parameters can be initialized by gradient-based optimization algorithms.
In addition, the sampling steps in \Cref{alg:sampling}
can be parallelized across layers and across training samples,
making it possible to sample the latent random noises by mini-batches.


\subsection{Predictive Density}
\label{sec:likelihood}

We further evaluate the properties of the predictive density function.
\Cref{thm:predictive-density} expresses the predictive density
in a recursive formulation.  Let $\phi_{\MVN}(\cdot;\bfmu,\bfSigma)$ be a multivariate normal density function with mean $\bfmu$ and covariance $\bfSigma$.

\begin{theorem}
  \label{thm:predictive-density}
  For $l \in \{ 0, \ldots, L\}$,
  let $\bftheta_{l} = \{\bfbeta_{l}, \bfgamma_{l}, \bfT_{l}, \bfSigma_{l}\}$
  and $\bfTheta_l = \{\bftheta_{l'}\}_{l'=0}^l$.
  In a B-DeepNoise model, the conditional density of the output $\bfy = \bfv_L$ given input value $\bfx$
  and model parameters $\bfTheta_l$
can  be iteratively constructed by
 \begin{align*}
  f(\bfv_{l} \mid \bfx, \bfTheta_{l})
  = &
  \int_{\bbR^{K_{l-1}}}
  \phi_{\MVN} \Big\{\bfv_{{l}} \Bigm| \bfgamma_{l} + \bfbeta_{l} h(\bfv_{l-1}), \bfT_l+\bfbeta_l\bfSigma_{l}\bfbeta_l^\top \Big\}
  f(\bfv_{l-1} \mid \bfx, \bfTheta_{l-1})
  d \bfv_{l-1},
\end{align*}
for $l\in\{1,\ldots, L\}$ and $ f(\bfv_{0} \mid \bfx, \bftheta_0)
  =
  \phi_{\MVN}(
    \bfv_{0} \mid
    \bfgamma_{0} + \bfbeta_{0}\bfx,
    \bfT_{0})$.
\end{theorem}

\begin{remark}
  As shown in \Cref{thm:predictive-density},
  the predictive distribution given an input value
  can be expressed as a continuous mixture of
  multivariate normal (CMMVN) distributions,
  where the mixing density is another CMMVN,
  depending on the intermediate values of the previous layer.
  Although this highly flexible predictive density
  does not have a closed-form expression,
  it can be simulated easily by adding normal noise
  to the intermediate values of the hidden layers,
  as stated in \Cref{eq:v,eq:u}.
\end{remark}

In standard DNNs and BNNs, 
predictive variance is completely determined
by the variance of the noise variable in the last layer.
As a more general model,
B-DeepNoise propagates variations in the latent random noises
to produce complex variation in the outcome.
When the variances of the latent random noises are all zero,
B-DeepNoise is reduced to a standard BNN.
A natural question is how the variance in the output variable
can be decomposed into variance of the latent random noises.
\Cref{thm:variance-bound-general} bounds the outcome variance
by the other model parameters.

\begin{theorem}
  \label{thm:variance-bound-general}
  Let $\bfy$ be the output value of the B-DeepNoise model
  given input value $\bfx$
  and model parameters
  $\bfTheta = \{\bfbeta_{l}, \bfgamma_{l}, \bfT_{l}, \bfSigma_{l}\}_{l=0}^L$.
  Let
  $g(\bfx, \bfGamma)$ be the output value of the standard DNN model
  with the same activation function and weight-bias parameters
  $\bfGamma = \{\bfbeta_{l}, \bfgamma_{l}\}_{l=0}^L$.
  (Note that $\bfy$ equals to $g(\bfx, \bfGamma)$ with probability one
  when all the latent noise variances are zero.)
  Assume the activation function $h$
  is Lipschitz continuous with Lipschitz constant $C_h$, 
  and define $d_l^2 = \|\bfbeta_l\|^2_2$.
  Then
  \begin{gather*}
    \Var(\bfy \mid \bfx, \bfTheta)
    +
    [ \E( \bfy \mid \bfx, \bfTheta)
    - g(\bfx, \bfGamma)]^2 
    \leq
    \sum_{l=0}^L
    \Bigg[  d_l^2 \sum_{k=1}^{K_{l-1}} \sigma^2_{l-1,k} + \sum_{k=1}^{K_{l}} \tau^2_{l,k} \Bigg]
    \Bigg[ \prod_{l'=l+1}^L d_{l'}^2 \Bigg]
    C^{2(L-l)}_{h}.
  \end{gather*} 
\end{theorem}

\begin{remark}
  According to \Cref{thm:variance-bound-general},
  given the model parameters,
  the predictive variance of B-DeepNoise
  is bounded by the latent noise variances, the spectrum norm of the weight matrices,
  and the Lipschitz constant of the activation function.
  In addition, the same upper bound holds for
  the squared distance of B-DeepNoise's predictive mean
  from the corresponding deterministic DNN's output value.
  The expression of this bound can be simplified
  for common activation functions and by using global bounds of model parameters,
  as shown in \Cref{thm:variance-bound-special}.
\end{remark}

\begin{corollary}
  \label{thm:variance-bound-special}
  Let
  $
    K = \underset{-1 \leq l \leq L}{\max} K_l
  $,
  $
    d^2 = \underset{0 \leq l \leq L}{\max} \| \bfbeta_l \|_2^2
  $,
  $
    \sigma^2 = \underset{0 \leq l \leq L-1}{\max} \| \bfsigma_l^2 \|_\infty
  $,
  $
    \tau^2 = \underset{0 \leq l \leq L}{\max} \| \bftau_l^2 \|_\infty
  $.
  Suppose activation function $h$ is
  sigmoid, tanh, hard sigmoid, hard tanh, ReLU, or leaky ReLU.
  Then
  \[
    \Var(\bfy \mid \bfx, \bfTheta)
    +
    [ \E( \bfy \mid \bfx, \bfTheta)
    - g(\bfx, \bfGamma)]^2 
    \leq
    KL
    \left(d^{2L} + 1\right)
    \left(d^2 + 1\right)
    \left(\sigma^2 + \tau^2\right).
  \]
\end{corollary}
For all the theorems in this work, see Supplementary Materials for proofs.

\section{Method Comparison}
We applied B-DeepNoise and existing methods to synthetic and real data
to evaluate their prediction accuracy,
predictive density estimation accuracy,
and uncertainty quantification efficiency.

\subsection{Experiments on Synthetic Data}
\label{sec:simulations}

\begin{table}[t]
  \scriptsize
  \captionsetup{font=footnotesize}
  \caption{
    Functional estimation error of the predictive distribution
    estimated by B-DeepNoise and baseline methods
    on synthetic datasets for various training sizes.
    Average performance is displayed in unit of 0.001.
    \label{tab:simu}
  }
  \centering
  \begin{tabular}{rrrrrrrrrr}
    \toprule
    \multicolumn{1}{c}{} & \multicolumn{3}{c}{heteroscedastic noise} & \multicolumn{3}{c}{skewed noise} & \multicolumn{3}{c}{multimodal noise} \\
    \cmidrule(lr){2-4} \cmidrule(lr){5-7} \cmidrule(lr){8-10}
    Method            & N=1000        &    N=2000     &     N=4000    &    N=1000     &     N=2000    &     N=4000    &  N=1000       &     N=2000    &     N=4000    \\
    \midrule
    BP                &     52 \mp 6  &     40 \mp 6  &     33 \mp 2  &    100 \mp 1  &     99 \mp 2  &     98 \mp 2  &     82 \mp 8  &     78 \mp 2  &     78 \mp 2  \\
    VI                &    103 \mp 1  &    103 \mp 1  &    101 \mp 1  &    103 \mp 2  &    102 \mp 1  &     98 \mp 1  &    173 \mp 2  &    175 \mp 1  &    175 \mp 0  \\
    BNN               &    103 \mp 2  &    102 \mp 2  &     95 \mp 2  &    105 \mp 2  &    104 \mp 2  &    101 \mp 1  &    128 \mp 8  &    128 \mp 6  &    122 \mp 6  \\
    DE                & \bf{49 \mp 4} &     39 \mp 4  &     32 \mp 2  &    101 \mp 1  &    100 \mp 3  &     97 \mp 1  & \bf{82 \mp 1} &     78 \mp 1  &     78 \mp 1  \\
    B-DeepNoise       &     72 \mp 4  & \bf{38 \mp 3} & \bf{26 \mp 2} & \bf{84 \mp 6} & \bf{53 \mp 7} & \bf{39 \mp 3} &     98 \mp 6  & \bf{60 \mp 4} & \bf{47 \mp 5} \\
    \bottomrule
  \end{tabular}
\end{table}

We used synthetic datasets to evaluate the predictive density
estimated by each method against the ground truth.
The input variable $x$ was one-dimensional and uniformly distributed
on $[-1, 1]$.
The output variable $y$ was also one-dimensional,
and its conditional median was a linear spline
with respect to $x$.
We designed the noise distribution
to be heteroscedastic, asymmetric, or multimodal,
as shown by the blue circles in \Cref{fig:simu}.
Training sample size varied among 1000, 2000, and 4000.
Every experimental setting was repeated for 20 times.

We used a B-DeepNoise model with 4 hidden layers
and 50 nodes in each layer,
with the hard tanh function as the activation function.
The prior distributions of the variance parameters were set to
$\InvGamma(0.001, 0.001)$.
We used gradient descent to initialize the model parameters
and drew 500 posterior samples.
B-DeepNoise was compared against
backpropagation (BP) with learnable predictive variance,
variational inference (VI)  \citep{ritter2022tyxe},
Bayesian neural networks (BNN)
with Hamiltonian Monte Carlo \citep{neal2011mcmc},
and deep ensemble (DE) \citep{lakshminarayanan2016simple}.
The baseline methods used identical architecture as B-DeepNoise,
and the hyperparameters were selected
according to the original authors' recommendations.
Details of the experiments are described in Supplementary Materials.

As visualized in \Cref{fig:simu},
B-DeepNoise successfully captured key characteristics
of the noise densities.
The estimated predictive distributions identified variation in
the output variance, regions with opposite directions of skewedness,
and abrupt changes from unimodal to bimodal distributions.
In contrast, predictive densities estimated by DE,
which is the best baseline method,
were mostly unimodal and symmetric
and overestimated the predictive variance overall.
Furthermore, the predictive densities estimated by VI
were primarily homoscedastic.

To quantitatively evaluate the accuracy
of the estimated predictive density functions,
we numerically computed the $L_1$ distance
between the inverse CDFs of the true and estimated
predictive distributions
over a grid of input values.
The average estimation errors are reported in \Cref{tab:simu}.
Among all the methods, B-DeepNoise had the smallest error
in all but two settings.
Moreover, for all three types of noises,
B-DeepNoise's accuracy improved much faster than the baseline methods
as the training sample size increased,
especially on the skewed and multimodal data.
These simulations illustrate the accuracy and efficiency of B-DeepNoise
in learning complex predictive density functions.

\subsection{Experiments on Real Data}
\label{sec:uci}

\begin{table}[t]
  \scriptsize
  \captionsetup{font=footnotesize}
  \centering
  \caption{
    UCI experiment results on testing data
    for B-DeepNoise and baseline methods.
    \label{tab:uci}
    }
  \begin{tabular}{lllllll}
    \toprule
    Dataset             & BP            & VI                 &  BNN           & DMC               & DE            & B-DeepNoise         \\
    \midrule
    \multicolumn{7}{c}{Root Mean Squared Error (RMSE)} \\
    \midrule                                                  
    Yacht Hydrodynamics &     3.09 \mp 1.28  &     2.10 \mp 0.43  &  0.68 \mp 0.04 &     3.30 \mp 1.14 &     2.02 \mp 0.76  & \bf{ 0.64 \mp 0.32} \\
    Boston Housing      &     3.36 \mp 1.05  & \bf{2.75 \mp 0.67} &  5.60 \mp 0.12 &     3.10 \mp 0.88 &     3.16 \mp 1.11  &      2.84 \mp 0.69  \\
    Energy Efficiency   &     2.45 \mp 0.32  &     0.68 \mp 0.09  & 10.31 \mp 0.16 &     1.46 \mp 0.18 &     2.56 \mp 0.32  & \bf{ 0.45 \mp 0.07} \\
    Concrete Strength   &     5.98 \mp 0.62  &     4.68 \mp 0.52  & 17.14 \mp 0.31 &     6.11 \mp 0.47 &     5.45 \mp 0.55  & \bf{ 4.54 \mp 0.46} \\
    Wine Quality        &     0.64 \mp 0.05  &     0.66 \mp 0.07  &  0.69 \mp 0.01 & \bf{0.62 \mp 0.04}& \bf{0.62 \mp 0.04} &      0.63 \mp 0.04  \\
    Kin8nm              &     0.08 \mp 0.00  &     0.09 \mp 0.00  &  0.16 \mp 0.00 &     2.27 \mp 0.23 & \bf{0.07 \mp 0.00} & \bf{ 0.07 \mp 0.00} \\
    Power Plant         &     4.02 \mp 0.15  &     3.89 \mp 0.20  & 17.74 \mp 0.10 &     4.12 \mp 0.15 &     3.98 \mp 0.15  & \bf{ 3.62 \mp 0.18} \\
    Naval Propulsion    & \bf{0.00 \mp 0.00} & \bf{0.00 \mp 0.00} &  0.02 \mp 0.00 &   509.93 \mp 0.00 & \bf{0.00 \mp 0.00} & \bf{ 0.00 \mp 0.00} \\
    Protein Structure   &     4.08 \mp 0.06  &     4.35 \mp 0.09  &  6.31 \mp 0.05 &     4.02 \mp 0.04 &     3.91 \mp 0.03  & \bf{ 3.64 \mp 0.03} \\
    \midrule
    \multicolumn{7}{c}{Negative Log Likelihood (NLL)} \\
    \midrule
    Yacht Hydrodynamics &  1.39 \mp 0.33 &  2.64 \mp 0.04 &  0.88 \mp 0.08 &      2.29 \mp 1.14  &      1.09 \mp 0.19  & \bf{ 0.45 \mp 0.22} \\
    Boston Housing      &  3.01 \mp 0.90 &  2.40 \mp 0.12 &  3.18 \mp 0.02 &      2.42 \mp 0.88  &      2.37 \mp 0.27  & \bf{ 2.28 \mp 0.17} \\
    Energy Efficiency   &  1.77 \mp 0.59 &  1.36 \mp 0.06 &  3.33 \mp 0.01 &      1.79 \mp 0.18  &      1.54 \mp 0.25  & \bf{ 0.55 \mp 0.19} \\
    Concrete Strength   &  3.41 \mp 0.47 &  3.03 \mp 0.21 &  4.24 \mp 0.01 &      3.19 \mp 0.47  &      3.04 \mp 0.22  & \bf{ 2.84 \mp 0.14} \\
    Wine Quality        &  2.07 \mp 0.86 & 10.05 \mp 4.03 &  1.02 \mp 0.01 & \bf{ 0.92 \mp 0.04} &      1.03 \mp 0.25  &      0.95 \mp 0.09  \\
    Kin8nm              & -1.01 \mp 0.21 &  1.67 \mp 0.44 & -0.40 \mp 0.03 &     -0.92 \mp 0.02  &     -1.30 \mp 0.04  & \bf{-1.31 \mp 0.04} \\
    Power Plant         &  2.81 \mp 0.06 &  2.92 \mp 0.10 &  4.18 \mp 0.01 &      2.80 \mp 0.03  &      2.77 \mp 0.05  & \bf{ 2.70 \mp 0.06} \\
    Naval Propulsion    & -4.67 \mp 1.29 & -7.19 \mp 0.48 & -2.09 \mp 0.03 &     -4.10 \mp 0.03  &     -5.15 \mp 0.21  & \bf{-7.25 \mp 0.07} \\
    Protein Structure   &  2.80 \mp 0.19 &  3.25 \mp 0.03 &  3.25 \mp 0.01 &      2.77 \mp 0.01  & \bf{ 2.54 \mp 0.05} &      2.65 \mp 0.09  \\
    \midrule
    \multicolumn{7}{c}{Width of 95\% Empirical Prediction Intervals (WEPI-95)} \\
    \midrule
    Yacht Hydrodynamics &  2.63 \mp 1.17     &  6.01 \mp 1.72     &  2.37 \mp 0.20 &    \inf \mp NA   &     2.34 \mp 0.91  & \bf{ 0.45 \mp 0.22} \\
    Boston Housing      &  \inf \mp NA       &  9.81 \mp 1.60     & 20.94 \mp 0.64 &   10.51 \mp 1.90 &     \inf \mp NA    & \bf{ 2.28 \mp 0.17} \\
    Energy Efficiency   &  3.70 \mp 1.05     &  2.60 \mp 0.46     & 16.79 \mp 0.54 &    6.28 \mp 1.14 &     3.55 \mp 0.79  & \bf{ 0.55 \mp 0.19} \\
    Concrete Strength   &  \inf \mp NA       &  \inf \mp NA       & 56.51 \mp 0.53 &   25.45 \mp 2.68 &    17.40 \mp 3.02  & \bf{ 2.84 \mp 0.14} \\
    Wine Quality        &  \inf \mp NA       &  \inf \mp NA       &  2.85 \mp 0.05 &    \inf \mp NA   &     2.88 \mp 0.34  & \bf{ 0.95 \mp 0.09} \\
    Kin8nm              &  0.30 \mp 0.02     &  \inf \mp NA       &  0.61 \mp 0.01 &   11.24 \mp 1.11 & \bf{0.26 \mp 0.01} & \bf{ 0.26 \mp 0.01} \\
    Power Plant         & 14.89 \mp 0.52     & 15.28 \mp 0.76     & 53.66 \mp 0.63 &    \inf \mp NA   &    14.76 \mp 0.56  & \bf{13.90 \mp 0.62} \\
    Naval Propulsion    &  \inf \mp NA       & \bf{0.00 \mp 0.00} &  0.05 \mp 0.00 & 2548.88 \mp 0.00 & \bf{0.00 \mp 0.00} & \bf{ 0.00 \mp 0.00} \\
    Protein Structure   & 16.32 \mp 0.63     &     \inf \mp NA    & 23.23 \mp 0.27 &    \inf \mp NA   &    14.06 \mp 0.37  & \bf{11.90 \mp 0.19} \\
    \bottomrule
  \end{tabular}
\end{table}

We applied B-DeepNoise and baselines methods to nine UCI regression datasets
\citep{dua2019uci}.
Experiment setup was similar to
\citep{gal2016dropout,lakshminarayanan2016simple},
where each dataset was randomly split into training
and testing sets for 20 times in all the datasets
except Protein Structure,
where only 5 random splits were made
due to the larger sample size.
Prediction accuracy was measured by
the testing root mean squared error (RMSE) of the predictive mean,
and the predictive distribution was assessed
by the testing negative log likelihood (NLL).
For efficiency of UQ, since the 95\% prediction intervals
might be miscalibrated, we computed
the 95\% \emph{empirical prediction intervals} (EPIs),
defined as the $x$\% prediction intervals
with minimal $x$ that cover
at least 95\% of the observed testing outcomes.
Then UQ efficiency was evaluated by the average width
of the 95\% EPIs (WEPI-95).
We compared B-DeepNoise against
backpropagation (BP) with learnable predictive variance,
variational inference (VI),
Bayesian Neural Network (BNN) with Hamiltonian Monte Carlo,
Dropout Monte Carlo (DMC),
and Deep Ensemble (DE).
All methods used 4 hidden layers with 50 nodes per layer,
with the hard tanh activation function.
Experiment setup is similar
to that of simulation studies.
(See Supplementary Materials for details.)

\Cref{tab:uci} shows the results on the testing data.
Compared to the baseline methods,
B-DeepNoise had the least RMSE on all except two datasets,
indicating a superior prediction accuracy.
Moreover, B-DeepNoise's predictive distribution
was also overall more accurate than the other methods,
since the NLL of B-DeepNoise was the smallest
on all except two datasets.
Furthermore, the high UQ efficiency of B-DeepNoise
was demonstrated by its uniformly narrowest WEPI-95.
In comparison,
the baseline methods not only had significantly larger WEPI-95s,
but also could not produce valid 95\% EPIs at all
on some of the datasets
(as indicated by $\infty$ in \Cref{tab:uci}),
since even their 100\% prediction intervals
could not cover at least 95\% of the testing outcomes.
In addition,
B-DeepNoise's performance was overall more stable than the baseline methods,
as reflected by its generally smaller standard errors of all the metrics.
In sum, the experiments on the UCI datasets demonstrated
the superior DR accuracy and UQ efficiency of B-DeepNoise
compared to existing methods.

\section{
  Neuroimaging-Based Prediction of General Intelligence for Adolescents
  }
\label{sec:abcd}

\begin{figure}[t]
  \scriptsize
  \captionsetup{font=footnotesize}
  \centering
  \caption{
    Observed g-scores (white crosses)
    and predictive densities (violin plots)
    estimated by B-DeepNoise
    for 19 selected testing subjects.
  }
  \begin{subfigure}[b]{0.48\textwidth}
    \caption{
      Imaging features included as predictors
      \label{fig:violin-withimg}
    }
    \includegraphics[width=\textwidth]{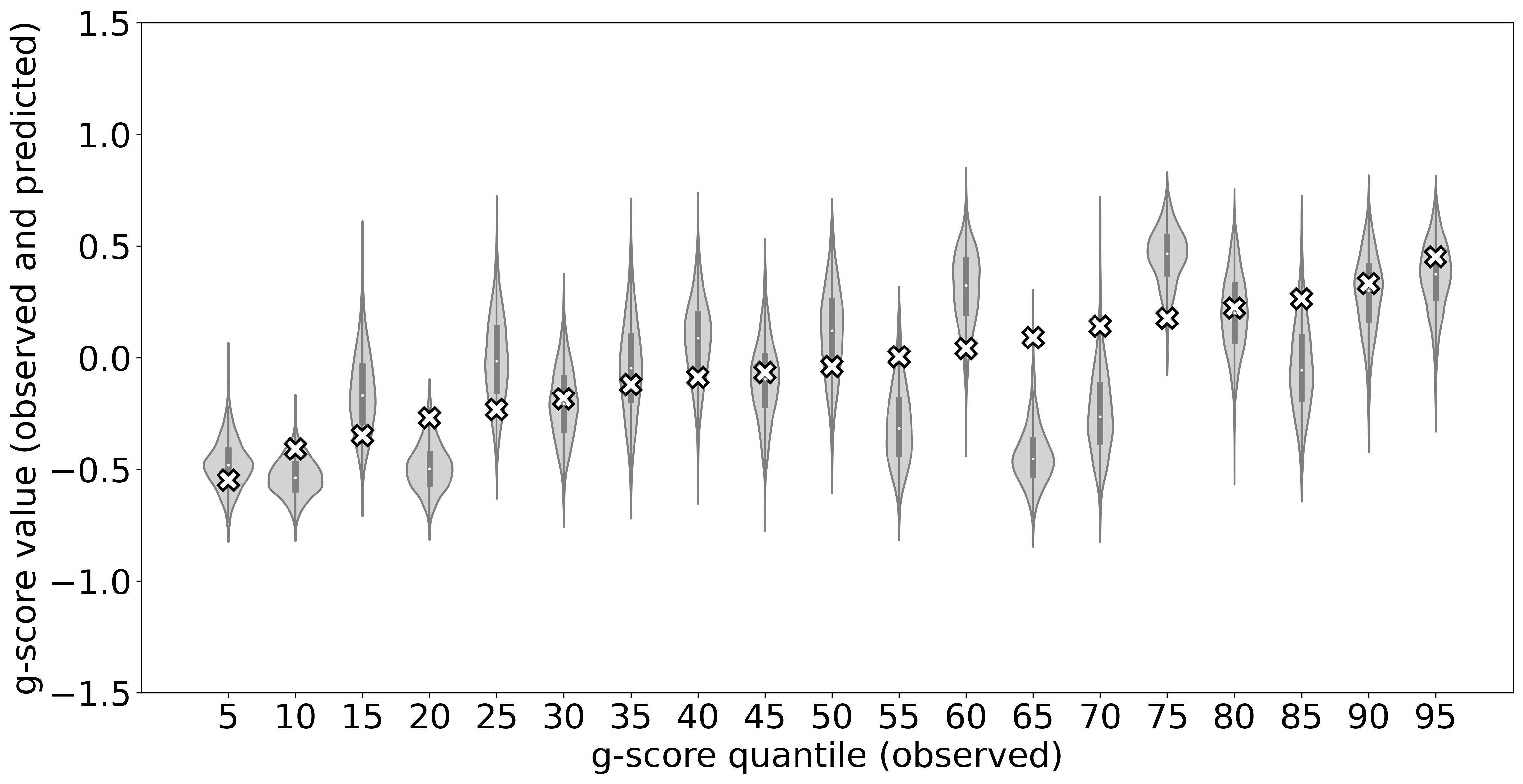}
  \end{subfigure}
  \hfill
  \begin{subfigure}[b]{0.48\textwidth}
    \caption{
      Imaging features excluded as predictors
      \label{fig:violin-noimg}
    }
    \includegraphics[width=\textwidth]{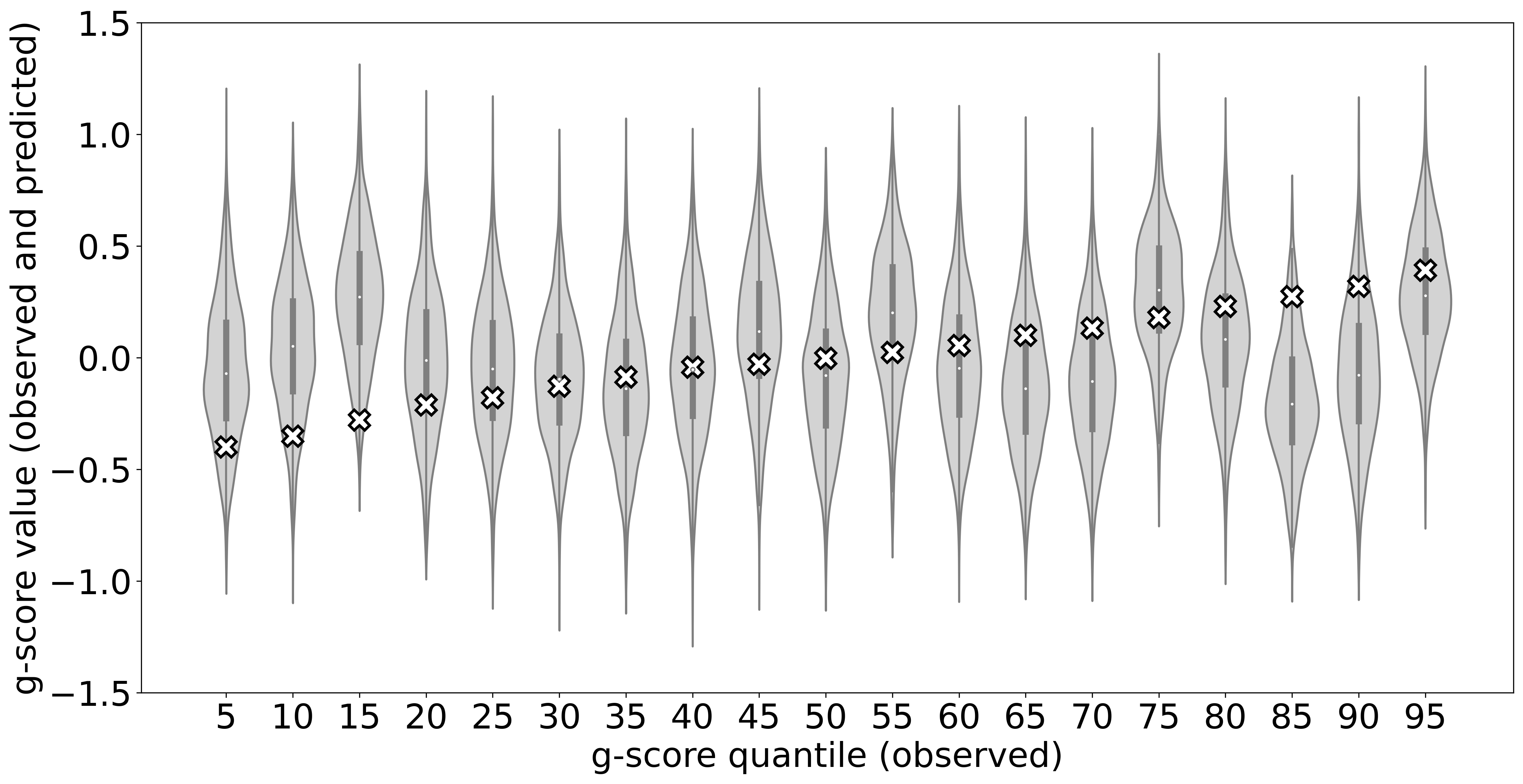}
  \end{subfigure}
\end{figure}

We demonstrate the usefulness of B-DeepNoise
in scientific studies by applying it to 
the neuroimaging data in the 
Adolescent Brain Cognitive Development (ABCD) Study \citep{casey2018adolescent,zhang2020image}.
The dataset contains 1191 subjects
aged between 9 and 10 years old,
recruited from multiple study sites in the United States.
For every subject, the data contain
general intelligence score (g-score) \citep{dubois2018distributed,o2016cognitive},
2-back task score \citep{cohen2016adolescent},
general psychopathology factor \citep{caspi2014p,carver2017toward,murray2016development},
demographic information
(including age, sex, parental education level,
household marital status, household income,
and ethnic backgrounds),
and brain functional magnetic resonance imagings (fMRIs).
In this study,
each subject is asked to participate in the 2-back task,
which is a functional assessment task
designed to engage memory regulation processes,
and a score is produced at the end to summarize the performance.
In addition, brain fMRI scans are acquired at resting state
and during the 2-back task.
The brain fMRI images in our dataset,
which have resolution $61 \times 73 \times 61$,
are obtained by contrasting
the resting state scans
and the task-active scans
with minimal processing.
These images reflect brain activity that is stimulated
by the 2-back task.

The goal of our analysis is to predict the g-score
by using the other features as predictors.
To this end, we train a B-DeepNoise model
by using the imaging and non-imaging features.
In order to provide biological interpretations,
we segmented the whole brain volume into 90 regions
according to the automated anatomical labeling (AAL) atlas \citep{tzourio2002automated}.
Then inside each region, we computed the mean of the imaging values
for each subject.
In total, we had 90 imaging predictors and 8 non-imaging predictors.
We randomly selected 90\% of the samples for training
and the rest for testing.
The hyperparameters for B-DeepNoise are similar to those
in the analysis of the UCI datasets
See Supplementary Materials for details of the experiment setup.

To illustrate the predictive density functions
learned by B-DeepNoise,
we selected 19 testing subjects
that correspond to the
5\%, 10\%, ..., 90\%, 95\% quantiles of the observed g-score.
The results are shown in \Cref{fig:violin-withimg}.
The prediction distributions estimated by B-DeepNoise have successfully covered
the observed outcomes, with only a couple of samples located near the tails
of the predictive distributions.
To assess B-DeepNoise's UQ efficiency,
we removed the imaging predictors
and refit the model with the non-imaging predictors only.
As shown in \Cref{fig:violin-noimg},
when the imaging information was not available,
B-DeepNoise widened the prediction intervals
to account for the higher degree of uncertainty.
In contrast, the predictive densities in the imaging-included model
are not only more concentrated but also exhibited greater
magnitude of heteroscedasticity and skewedness.
These results indicate that B-DeepNoise is able to appropriately adjust
the predictive density to reflect its individualized degree of confidence
in its predictions.

\begin{figure}[t]
  \footnotesize
  \captionsetup{font=footnotesize}
  \centering
  \caption{
    Influence of brain regions on the predictive mean
    of the g-score.
    \label{fig:brain}
  }
  \includegraphics[height=1in]{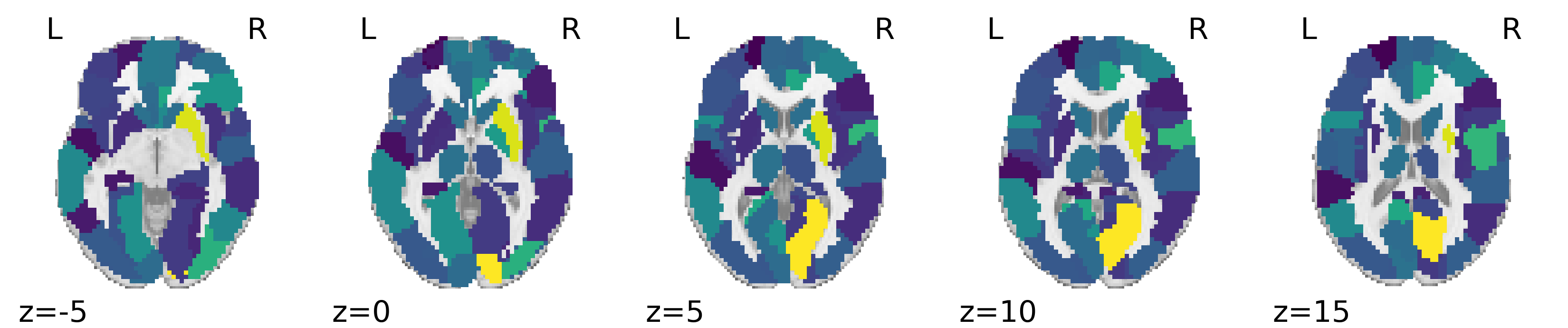}
  \includegraphics[height=1in]{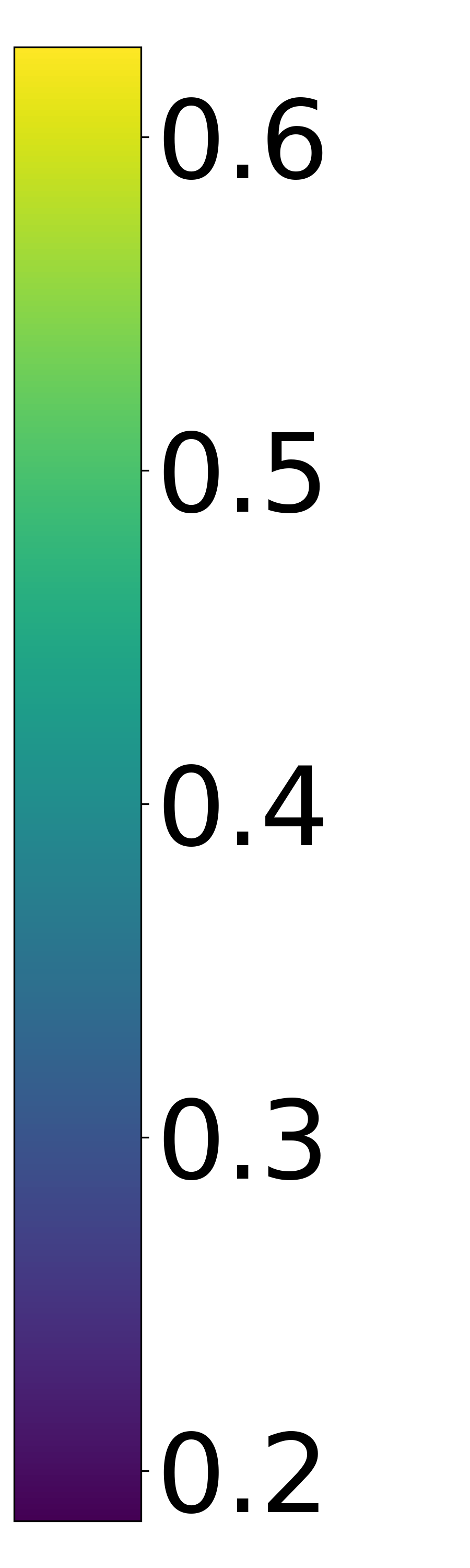}
\end{figure}

Furthermore, we investigated the most influential features
on the predictive mean of the g-score.
Then influence of a feature on the predictive mean
is measured by the average absolute value of the gradient
of the predictive mean with respect to the feature
in the B-DeepNoise model.
First of all, the 2-back task score had
the highest influence ($1.32$) on the predicted mean g-score,
and the magnitude of influence was much higher than the
other features ($< 0.63$, see Supplementary Materials).
This result is consistent with the current understanding that
memory is a one of the major components that encompass cognitive abilities
\citep{thompson2019structure}.
The rest of the most influential features
were primarily imaging features
(see \Cref{fig:brain} for the influence of brain regions on predicted g-scores),
and many of the corresponding brain regions have been well studied
for their associations with general intelligence in existing studies.
To name a few,
bilateral calcarine is associated with
the intelligence quotient (IQ)
of children and adolescents \citep{kilroy2011relationships};
putamen has been identified with verbal IQ
in healthy adults \citep{grazioplene2015subcortical};
the right paracentral lobule
has been found to be associated with
functioning decline in at-risk mental state patients
\citep{sasabayashi2021reduced}.
Overall, neuroimaging regions that are the most influential
for B-DeepNoise's predicted mean g-score
are supported by existing findings in the literature.

\section{Discussion}
\label{sec:discussion}

In this work, we have presented B-DeepNoise,
a Bayesian nonparametric model capable of DR and UQ.
Intuitively, we generalize standard DNNs and BNNs
by extending the random noise variables
from the output layer to all the hidden layers.
By iteratively composing latent random noise
with linear and nonlinear transformations in the network,
B-DeepNoise can learn highly complex
predictive density functions.
Moreover, the addition of the latent noise
leads to closed-form full conditional posterior distributions
of the model parameters,
which we exploit to develop a Gibbs sampling algorithm
for simulating the posterior distribution
without relying on computationally intensive MH steps.
Furthermore, we have derived theoretical properties of B-DeepNoise
regarding its predictive density and variance propagation.
Simulations have demonstrated B-DeepNoise's capacity
in approximating heteroscedastic, asymmetric, and multimodal
predictive densities.
As shown by the analysis of real benchmark datasets,
compared to existing methods,
B-DeepNoise not only has more accurate predictive means
and predictive densities
but is also more efficient in uncertainty quantification.
Finally, we applied B-DeepNoise to predict
genera intelligence by using neuroimaging and non-imaging
in the ABCD study.

In this work, we have focused on
estimating the conditional predictive densities
of the output given the input,
but we have not considered the marginal densities of the input.
In the future, we plan on including the marginal distributions
as a part of our DR and UQ model.
Moreover, we are also interested in developing robust frequentist methods
for the B-DeepNoise model,
as the current method focuses on Bayesian frameworks.
Furthermore, the current version of B-DeepNoise is limited to
continuous outcomes.
For categorical outcomes, a softmax function can be appended
to the output layer and approximated by piecewise linear functions.
Then the Gibbs sampler for B-DeepNoise can be easily extended
to categorical outcome variables.
See Supplementary Materials for a more detailed discussion.
In addition, we have yet investigated B-DeepNoise on multi-outcome data.
Our model has the potential as an outcome selection method,
where the goal is to differentiate
predictable and unpredictable outcome variables.

\newpage
\bibliographystyle{unsrt}
\bibliography{main}

\begin{thebibliography}{10}

\bibitem{goodfellow2016deep}
Ian Goodfellow, Yoshua Bengio, Aaron Courville, and Yoshua Bengio.
\newblock {\em Deep learning}, volume~1.
\newblock MIT press Cambridge, 2016.

\bibitem{lecun2015deep}
Yann LeCun, Yoshua Bengio, and Geoffrey Hinton.
\newblock Deep learning.
\newblock {\em nature}, 521(7553):436--444, 2015.

\bibitem{pouyanfar2018survey}
Samira Pouyanfar, Saad Sadiq, Yilin Yan, Haiman Tian, Yudong Tao, Maria~Presa
  Reyes, Mei-Ling Shyu, Shu-Ching Chen, and Sundaraja~S Iyengar.
\newblock A survey on deep learning: Algorithms, techniques, and applications.
\newblock {\em ACM Computing Surveys (CSUR)}, 51(5):1--36, 2018.

\bibitem{berner2021modern}
Julius Berner, Philipp Grohs, Gitta Kutyniok, and Philipp Petersen.
\newblock The modern mathematics of deep learning.
\newblock {\em arXiv preprint arXiv:2105.04026}, 2021.

\bibitem{bojarski2016end}
Mariusz Bojarski, Davide Del~Testa, Daniel Dworakowski, Bernhard Firner, Beat
  Flepp, Prasoon Goyal, Lawrence~D Jackel, Mathew Monfort, Urs Muller, Jiakai
  Zhang, et~al.
\newblock End to end learning for self-driving cars.
\newblock {\em arXiv preprint arXiv:1604.07316}, 2016.

\bibitem{grigorescu2020survey}
Sorin Grigorescu, Bogdan Trasnea, Tiberiu Cocias, and Gigel Macesanu.
\newblock A survey of deep learning techniques for autonomous driving.
\newblock {\em Journal of Field Robotics}, 37(3):362--386, 2020.

\bibitem{ker2017deep}
Justin Ker, Lipo Wang, Jai Rao, and Tchoyoson Lim.
\newblock Deep learning applications in medical image analysis.
\newblock {\em Ieee Access}, 6:9375--9389, 2017.

\bibitem{zou2019primer}
James Zou, Mikael Huss, Abubakar Abid, Pejman Mohammadi, Ali Torkamani, and
  Amalio Telenti.
\newblock A primer on deep learning in genomics.
\newblock {\em Nature genetics}, 51(1):12--18, 2019.

\bibitem{stephenson2019survey}
Natalie Stephenson, Emily Shane, Jessica Chase, Jason Rowland, David Ries,
  Nicola Justice, Jie Zhang, Leong Chan, and Renzhi Cao.
\newblock Survey of machine learning techniques in drug discovery.
\newblock {\em Current drug metabolism}, 20(3):185--193, 2019.

\bibitem{zemouri2019deep}
Ryad Zemouri, Noureddine Zerhouni, and Daniel Racoceanu.
\newblock Deep learning in the biomedical applications: Recent and future
  status.
\newblock {\em Applied Sciences}, 9(8):1526, 2019.

\bibitem{begoli2019need}
Edmon Begoli, Tanmoy Bhattacharya, and Dimitri Kusnezov.
\newblock The need for uncertainty quantification in machine-assisted medical
  decision making.
\newblock {\em Nature Machine Intelligence}, 1(1):20--23, 2019.

\bibitem{amodei2016concrete}
Dario Amodei, Chris Olah, Jacob Steinhardt, Paul Christiano, John Schulman, and
  Dan Man{\'e}.
\newblock Concrete problems in ai safety.
\newblock {\em arXiv preprint arXiv:1606.06565}, 2016.

\bibitem{jiang2012calibrating}
Xiaoqian Jiang, Melanie Osl, Jihoon Kim, and Lucila Ohno-Machado.
\newblock Calibrating predictive model estimates to support personalized
  medicine.
\newblock {\em Journal of the American Medical Informatics Association},
  19(2):263--274, 2012.

\bibitem{leibig2017leveraging}
Christian Leibig, Vaneeda Allken, Murat~Se{\c{c}}kin Ayhan, Philipp Berens, and
  Siegfried Wahl.
\newblock Leveraging uncertainty information from deep neural networks for
  disease detection.
\newblock {\em Scientific reports}, 7(1):1--14, 2017.

\bibitem{dunson2007bayesian}
David~B Dunson, Natesh Pillai, and Ju-Hyun Park.
\newblock Bayesian density regression.
\newblock {\em Journal of the Royal Statistical Society: Series B (Statistical
  Methodology)}, 69(2):163--183, 2007.

\bibitem{abdar2021review}
Moloud Abdar, Farhad Pourpanah, Sadiq Hussain, Dana Rezazadegan, Li~Liu,
  Mohammad Ghavamzadeh, Paul Fieguth, Xiaochun Cao, Abbas Khosravi, U~Rajendra
  Acharya, et~al.
\newblock A review of uncertainty quantification in deep learning: Techniques,
  applications and challenges.
\newblock {\em Information Fusion}, 76:243--297, 2021.

\bibitem{staahl2020evaluation}
Niclas St{\aa}hl, G{\"o}ran Falkman, Alexander Karlsson, and Gunnar Mathiason.
\newblock Evaluation of uncertainty quantification in deep learning.
\newblock In {\em International Conference on Information Processing and
  Management of Uncertainty in Knowledge-Based Systems}, pages 556--568.
  Springer, 2020.

\bibitem{caldeira2020deeply}
Jo{\~a}o Caldeira and Brian Nord.
\newblock Deeply uncertain: comparing methods of uncertainty quantification in
  deep learning algorithms.
\newblock {\em Machine Learning: Science and Technology}, 2(1):015002, 2020.

\bibitem{zhu2019physics}
Yinhao Zhu, Nicholas Zabaras, Phaedon-Stelios Koutsourelakis, and Paris
  Perdikaris.
\newblock Physics-constrained deep learning for high-dimensional surrogate
  modeling and uncertainty quantification without labeled data.
\newblock {\em Journal of Computational Physics}, 394:56--81, 2019.

\bibitem{bishop1994mixture}
Christopher~M Bishop.
\newblock Mixture density networks.
\newblock 1994.

\bibitem{bishop2006pattern}
Christopher~M Bishop and Nasser~M Nasrabadi.
\newblock {\em Pattern recognition and machine learning}, volume~4.
\newblock Springer, 2006.

\bibitem{lakshminarayanan2016simple}
Balaji Lakshminarayanan, Alexander Pritzel, and Charles Blundell.
\newblock Simple and scalable predictive uncertainty estimation using deep
  ensembles.
\newblock {\em arXiv preprint arXiv:1612.01474}, 2016.

\bibitem{lei2014distribution}
Jing Lei and Larry Wasserman.
\newblock Distribution-free prediction bands for non-parametric regression.
\newblock {\em Journal of the Royal Statistical Society: Series B (Statistical
  Methodology)}, 76(1):71--96, 2014.

\bibitem{pearce2018high}
Tim Pearce, Alexandra Brintrup, Mohamed Zaki, and Andy Neely.
\newblock High-quality prediction intervals for deep learning: A
  distribution-free, ensembled approach.
\newblock In {\em International conference on machine learning}, pages
  4075--4084. PMLR, 2018.

\bibitem{romano2019conformalized}
Yaniv Romano, Evan Patterson, and Emmanuel Candes.
\newblock Conformalized quantile regression.
\newblock {\em Advances in neural information processing systems}, 32, 2019.

\bibitem{tagasovska2019single}
Natasa Tagasovska and David Lopez-Paz.
\newblock Single-model uncertainties for deep learning.
\newblock {\em Advances in Neural Information Processing Systems}, 32, 2019.

\bibitem{li2021deep}
Rui Li, Brian~J Reich, and Howard~D Bondell.
\newblock Deep distribution regression.
\newblock {\em Computational Statistics \& Data Analysis}, 159:107203, 2021.

\bibitem{huberman2021nonparametric}
David~B Huberman, Brian~J Reich, and Howard~D Bondell.
\newblock Nonparametric conditional density estimation in a deep learning
  framework for short-term forecasting.
\newblock {\em Environmental and Ecological Statistics}, pages 1--15, 2021.

\bibitem{mackay1995probable}
David~JC MacKay.
\newblock Probable networks and plausible predictions-a review of practical
  bayesian methods for supervised neural networks.
\newblock {\em Network: computation in neural systems}, 6(3):469, 1995.

\bibitem{neal2012bayesian}
Radford~M Neal.
\newblock {\em Bayesian learning for neural networks}, volume 118.
\newblock Springer Science \& Business Media, 2012.

\bibitem{xue2019reliable}
Yujia Xue, Shiyi Cheng, Yunzhe Li, and Lei Tian.
\newblock Reliable deep-learning-based phase imaging with uncertainty
  quantification.
\newblock {\em Optica}, 6(5):618--629, 2019.

\bibitem{hwang1997prediction}
JT~Gene Hwang and A~Adam Ding.
\newblock Prediction intervals for artificial neural networks.
\newblock {\em Journal of the American Statistical Association},
  92(438):748--757, 1997.

\bibitem{wang2020uncertainty}
Yuexi Wang and Veronika Rockov{\'a}.
\newblock Uncertainty quantification for sparse deep learning.
\newblock In {\em International Conference on Artificial Intelligence and
  Statistics}, pages 298--308. PMLR, 2020.

\bibitem{sun2021sparse}
Yan Sun, Wenjun Xiong, and Faming Liang.
\newblock Sparse deep learning: A new framework immune to local traps and
  miscalibration.
\newblock {\em Advances in Neural Information Processing Systems}, 34, 2021.

\bibitem{kendall2017uncertainties}
Alex Kendall and Yarin Gal.
\newblock What uncertainties do we need in bayesian deep learning for computer
  vision?
\newblock {\em arXiv preprint arXiv:1703.04977}, 2017.

\bibitem{izmailov2018averaging}
Pavel Izmailov, Dmitrii Podoprikhin, Timur Garipov, Dmitry Vetrov, and
  Andrew~Gordon Wilson.
\newblock Averaging weights leads to wider optima and better generalization.
\newblock {\em arXiv preprint arXiv:1803.05407}, 2018.

\bibitem{blei2017variational}
David~M Blei, Alp Kucukelbir, and Jon~D McAuliffe.
\newblock Variational inference: A review for statisticians.
\newblock {\em Journal of the American statistical Association},
  112(518):859--877, 2017.

\bibitem{kingma2013auto}
Diederik~P Kingma and Max Welling.
\newblock Auto-encoding variational bayes.
\newblock {\em arXiv preprint arXiv:1312.6114}, 2013.

\bibitem{mandt2017stochastic}
Stephan Mandt, Matthew~D Hoffman, and David~M Blei.
\newblock Stochastic gradient descent as approximate bayesian inference.
\newblock {\em Journal of Machine Learning Research}, 18:1--35, 2017.

\bibitem{graves2011practical}
Alex Graves.
\newblock Practical variational inference for neural networks.
\newblock {\em Advances in neural information processing systems}, 24, 2011.

\bibitem{louizos2016structured}
Christos Louizos and Max Welling.
\newblock Structured and efficient variational deep learning with matrix
  gaussian posteriors.
\newblock In {\em International conference on machine learning}, pages
  1708--1716. PMLR, 2016.

\bibitem{lee2020estimating}
Jongseok Lee, Matthias Humt, Jianxiang Feng, and Rudolph Triebel.
\newblock Estimating model uncertainty of neural networks in sparse information
  form.
\newblock In {\em International Conference on Machine Learning}, pages
  5702--5713. PMLR, 2020.

\bibitem{rezende2015variational}
Danilo Rezende and Shakir Mohamed.
\newblock Variational inference with normalizing flows.
\newblock In {\em International conference on machine learning}, pages
  1530--1538. PMLR, 2015.

\bibitem{louizos2017multiplicative}
Christos Louizos and Max Welling.
\newblock Multiplicative normalizing flows for variational bayesian neural
  networks.
\newblock In {\em International Conference on Machine Learning}, pages
  2218--2227. PMLR, 2017.

\bibitem{srivastava2014dropout}
Nitish Srivastava, Geoffrey Hinton, Alex Krizhevsky, Ilya Sutskever, and Ruslan
  Salakhutdinov.
\newblock Dropout: a simple way to prevent neural networks from overfitting.
\newblock {\em The journal of machine learning research}, 15(1):1929--1958,
  2014.

\bibitem{gal2016dropout}
Yarin Gal and Zoubin Ghahramani.
\newblock Dropout as a bayesian approximation: Representing model uncertainty
  in deep learning.
\newblock In {\em international conference on machine learning}, pages
  1050--1059. PMLR, 2016.

\bibitem{molchanov2017variational}
Dmitry Molchanov, Arsenii Ashukha, and Dmitry Vetrov.
\newblock Variational dropout sparsifies deep neural networks.
\newblock In {\em International Conference on Machine Learning}, pages
  2498--2507. PMLR, 2017.

\bibitem{ioffe2015batch}
Sergey Ioffe and Christian Szegedy.
\newblock Batch normalization: Accelerating deep network training by reducing
  internal covariate shift.
\newblock In {\em International conference on machine learning}, pages
  448--456. PMLR, 2015.

\bibitem{teye2018bayesian}
Mattias Teye, Hossein Azizpour, and Kevin Smith.
\newblock Bayesian uncertainty estimation for batch normalized deep networks.
\newblock In {\em International Conference on Machine Learning}, pages
  4907--4916. PMLR, 2018.

\bibitem{blundell2015weight}
Charles Blundell, Julien Cornebise, Koray Kavukcuoglu, and Daan Wierstra.
\newblock Weight uncertainty in neural network.
\newblock In {\em International Conference on Machine Learning}, pages
  1613--1622. PMLR, 2015.

\bibitem{hernandez2015probabilistic}
Jos{\'e}~Miguel Hern{\'a}ndez-Lobato and Ryan Adams.
\newblock Probabilistic backpropagation for scalable learning of bayesian
  neural networks.
\newblock In {\em International conference on machine learning}, pages
  1861--1869. PMLR, 2015.

\bibitem{chib1995understanding}
Siddhartha Chib and Edward Greenberg.
\newblock Understanding the metropolis-hastings algorithm.
\newblock {\em The american statistician}, 49(4):327--335, 1995.

\bibitem{hitchcock2003history}
David~B Hitchcock.
\newblock A history of the metropolis--hastings algorithm.
\newblock {\em The American Statistician}, 57(4):254--257, 2003.

\bibitem{andrieu2008tutorial}
Christophe Andrieu and Johannes Thoms.
\newblock A tutorial on adaptive mcmc.
\newblock {\em Statistics and computing}, 18(4):343--373, 2008.

\bibitem{wenzel2020good}
Florian Wenzel, Kevin Roth, Bastiaan~S Veeling, Jakub {\'S}wi{\k{a}}tkowski,
  Linh Tran, Stephan Mandt, Jasper Snoek, Tim Salimans, Rodolphe Jenatton, and
  Sebastian Nowozin.
\newblock How good is the bayes posterior in deep neural networks really?
\newblock {\em arXiv preprint arXiv:2002.02405}, 2020.

\bibitem{wilson2020bayesian}
Andrew~Gordon Wilson and Pavel Izmailov.
\newblock Bayesian deep learning and a probabilistic perspective of
  generalization.
\newblock {\em arXiv preprint arXiv:2002.08791}, 2020.

\bibitem{welling2011bayesian}
Max Welling and Yee~W Teh.
\newblock Bayesian learning via stochastic gradient langevin dynamics.
\newblock In {\em Proceedings of the 28th international conference on machine
  learning (ICML-11)}, pages 681--688. Citeseer, 2011.

\bibitem{chen2014stochastic}
Tianqi Chen, Emily Fox, and Carlos Guestrin.
\newblock Stochastic gradient hamiltonian monte carlo.
\newblock In {\em International conference on machine learning}, pages
  1683--1691. PMLR, 2014.

\bibitem{chen2016bridging}
Changyou Chen, David Carlson, Zhe Gan, Chunyuan Li, and Lawrence Carin.
\newblock Bridging the gap between stochastic gradient mcmc and stochastic
  optimization.
\newblock In {\em Artificial Intelligence and Statistics}, pages 1051--1060.
  PMLR, 2016.

\bibitem{wu2020mini}
Tung-Yu Wu, YX~Rachel~Wang, and Wing~H Wong.
\newblock Mini-batch metropolis--hastings with reversible sgld proposal.
\newblock {\em Journal of the American Statistical Association}, pages 1--9,
  2020.

\bibitem{liang2016bootstrap}
Faming Liang, Jinsu Kim, and Qifan Song.
\newblock A bootstrap metropolis--hastings algorithm for bayesian analysis of
  big data.
\newblock {\em Technometrics}, 58(3):304--318, 2016.

\bibitem{jospin2022hands}
Laurent~Valentin Jospin, Hamid Laga, Farid Boussaid, Wray Buntine, and Mohammed
  Bennamoun.
\newblock Hands-on bayesian neural networks—a tutorial for deep learning
  users.
\newblock {\em IEEE Computational Intelligence Magazine}, 17(2):29--48, 2022.

\bibitem{geman1984stochastic}
Stuart Geman and Donald Geman.
\newblock Stochastic relaxation, gibbs distributions, and the bayesian
  restoration of images.
\newblock {\em IEEE Transactions on pattern analysis and machine intelligence},
  (6):721--741, 1984.

\bibitem{gelfand1990sampling}
Alan~E Gelfand and Adrian~FM Smith.
\newblock Sampling-based approaches to calculating marginal densities.
\newblock {\em Journal of the American statistical association},
  85(410):398--409, 1990.

\bibitem{roberts1994simple}
Gareth~O Roberts and Adrian~FM Smith.
\newblock Simple conditions for the convergence of the gibbs sampler and
  metropolis-hastings algorithms.
\newblock {\em Stochastic processes and their applications}, 49(2):207--216,
  1994.

\bibitem{gelfand2000gibbs}
Alan~E Gelfand.
\newblock Gibbs sampling.
\newblock {\em Journal of the American statistical Association},
  95(452):1300--1304, 2000.

\bibitem{you2019adversarial}
Zhonghui You, Jinmian Ye, Kunming Li, Zenglin Xu, and Ping Wang.
\newblock Adversarial noise layer: Regularize neural network by adding noise.
\newblock In {\em 2019 IEEE International Conference on Image Processing
  (ICIP)}, pages 909--913. IEEE, 2019.

\bibitem{gulcehre2016noisy}
Caglar Gulcehre, Marcin Moczulski, Misha Denil, and Yoshua Bengio.
\newblock Noisy activation functions.
\newblock In {\em International conference on machine learning}, pages
  3059--3068. PMLR, 2016.

\bibitem{lee2019probact}
Joonho Lee, Kumar Shridhar, Hideaki Hayashi, Brian~Kenji Iwana, Seokjun Kang,
  and Seiichi Uchida.
\newblock Probact: A probabilistic activation function for deep neural
  networks.
\newblock {\em arXiv preprint arXiv:1905.10761}, 5:13, 2019.

\bibitem{sun2022kernel}
Yan Sun and Faming Liang.
\newblock A kernel-expanded stochastic neural network.
\newblock {\em Journal of the Royal Statistical Society: Series B (Statistical
  Methodology)}, n/a(n/a), 2022.

\bibitem{plataniotis2017gaussian}
Kostantinos~N Plataniotis and Dimitris Hatzinakos.
\newblock Gaussian mixtures and their applications to signal processing.
\newblock In {\em Advanced signal processing handbook}, pages 89--124. CRC
  Press, 2017.

\bibitem{calcaterra2008approximating}
Craig Calcaterra and Axel Boldt.
\newblock Approximating with gaussians.
\newblock {\em arXiv preprint arXiv:0805.3795}, 2008.

\bibitem{scarselli1998universal}
Franco Scarselli and Ah~Chung Tsoi.
\newblock Universal approximation using feedforward neural networks: A survey
  of some existing methods, and some new results.
\newblock {\em Neural networks}, 11(1):15--37, 1998.

\bibitem{yarotsky2017error}
Dmitry Yarotsky.
\newblock Error bounds for approximations with deep relu networks.
\newblock {\em Neural Networks}, 94:103--114, 2017.

\bibitem{lu2020universal}
Yulong Lu and Jianfeng Lu.
\newblock A universal approximation theorem of deep neural networks for
  expressing probability distributions.
\newblock {\em Advances in Neural Information Processing Systems}, 33, 2020.

\bibitem{maas2013rectifier}
Andrew~L Maas, Awni~Y Hannun, Andrew~Y Ng, et~al.
\newblock Rectifier nonlinearities improve neural network acoustic models.
\newblock In {\em Proc. icml}, volume~30, page~3. Citeseer, 2013.

\bibitem{ritter2022tyxe}
Hippolyt Ritter and Theofanis Karaletsos.
\newblock Tyxe: Pyro-based bayesian neural nets for pytorch.
\newblock {\em Proceedings of Machine Learning and Systems}, 4, 2022.

\bibitem{neal2011mcmc}
Radford~M Neal et~al.
\newblock Mcmc using hamiltonian dynamics.
\newblock {\em Handbook of markov chain monte carlo}, 2(11):2, 2011.

\bibitem{dua2019uci}
Dheeru Dua and Casey Graff.
\newblock {UCI} machine learning repository, 2017.

\bibitem{casey2018adolescent}
BJ~Casey, Tariq Cannonier, May~I Conley, Alexandra~O Cohen, Deanna~M Barch,
  Mary~M Heitzeg, Mary~E Soules, Theresa Teslovich, Danielle~V Dellarco, Hugh
  Garavan, et~al.
\newblock The adolescent brain cognitive development (abcd) study: imaging
  acquisition across 21 sites.
\newblock {\em Developmental cognitive neuroscience}, 32:43--54, 2018.

\bibitem{zhang2020image}
Daiwei Zhang, Lexin Li, Chandra Sripada, and Jian Kang.
\newblock Image response regression via deep neural networks.
\newblock {\em arXiv preprint arXiv:2006.09911}, 2020.

\bibitem{dubois2018distributed}
Julien Dubois, Paola Galdi, Lynn~K Paul, and Ralph Adolphs.
\newblock A distributed brain network predicts general intelligence from
  resting-state human neuroimaging data.
\newblock {\em Philosophical Transactions of the Royal Society B: Biological
  Sciences}, 373(1756):20170284, 2018.

\bibitem{o2016cognitive}
Andrew O'Shea, Ronald Cohen, Eric~C Porges, Nicole~R Nissim, and Adam~J Woods.
\newblock Cognitive aging and the hippocampus in older adults.
\newblock {\em Frontiers in aging neuroscience}, 8:298, 2016.

\bibitem{cohen2016adolescent}
Alexandra~O Cohen, Kaitlyn Breiner, Laurence Steinberg, Richard~J Bonnie,
  Elizabeth~S Scott, Kim Taylor-Thompson, Marc~D Rudolph, Jason Chein,
  Jennifer~A Richeson, Aaron~S Heller, et~al.
\newblock When is an adolescent an adult? assessing cognitive control in
  emotional and nonemotional contexts.
\newblock {\em Psychological Science}, 27(4):549--562, 2016.

\bibitem{caspi2014p}
Avshalom Caspi, Renate~M Houts, Daniel~W Belsky, Sidra~J Goldman-Mellor,
  HonaLee Harrington, Salomon Israel, Madeline~H Meier, Sandhya Ramrakha, Idan
  Shalev, Richie Poulton, et~al.
\newblock The p factor: one general psychopathology factor in the structure of
  psychiatric disorders?
\newblock {\em Clinical psychological science}, 2(2):119--137, 2014.

\bibitem{carver2017toward}
Charles~S Carver, Sheri~L Johnson, and Kiara~R Timpano.
\newblock Toward a functional view of the p factor in psychopathology.
\newblock {\em Clinical Psychological Science}, 5(5):880--889, 2017.

\bibitem{murray2016development}
Aja~Louise Murray, Manuel Eisner, and Denis Ribeaud.
\newblock The development of the general factor of psychopathology ‘p
  factor’through childhood and adolescence.
\newblock {\em Journal of abnormal child psychology}, 44(8):1573--1586, 2016.

\bibitem{tzourio2002automated}
Nathalie Tzourio-Mazoyer, Brigitte Landeau, Dimitri Papathanassiou, Fabrice
  Crivello, Olivier Etard, Nicolas Delcroix, Bernard Mazoyer, and Marc Joliot.
\newblock Automated anatomical labeling of activations in spm using a
  macroscopic anatomical parcellation of the mni mri single-subject brain.
\newblock {\em Neuroimage}, 15(1):273--289, 2002.

\bibitem{thompson2019structure}
Wesley~K Thompson, Deanna~M Barch, James~M Bjork, Raul Gonzalez, Bonnie~J
  Nagel, Sara~Jo Nixon, and Monica Luciana.
\newblock The structure of cognition in 9 and 10 year-old children and
  associations with problem behaviors: Findings from the abcd study’s
  baseline neurocognitive battery.
\newblock {\em Developmental cognitive neuroscience}, 36:100606, 2019.

\bibitem{kilroy2011relationships}
Emily Kilroy, Collin~Y Liu, Lirong Yan, Yoon~Chun Kim, Mirella Dapretto,
  Mario~F Mendez, and Danny~JJ Wang.
\newblock Relationships between cerebral blood flow and iq in typically
  developing children and adolescents.
\newblock {\em Journal of cognitive science}, 12(2):151, 2011.

\bibitem{grazioplene2015subcortical}
Rachael~G Grazioplene, Sephira G.~Ryman, Jeremy~R Gray, Aldo Rustichini, Rex~E
  Jung, and Colin~G DeYoung.
\newblock Subcortical intelligence: Caudate volume predicts iq in healthy
  adults.
\newblock {\em Human brain mapping}, 36(4):1407--1416, 2015.

\bibitem{sasabayashi2021reduced}
Daiki Sasabayashi, Yoichiro Takayanagi, Tsutomu Takahashi, Shimako Nishiyama,
  Yuko Mizukami, Naoyuki Katagiri, Naohisa Tsujino, Takahiro Nemoto, Atsushi
  Sakuma, Masahiro Katsura, et~al.
\newblock Reduced cortical thickness of the paracentral lobule in at-risk
  mental state individuals with poor 1-year functional outcomes.
\newblock {\em Translational psychiatry}, 11(1):1--9, 2021.

\end{thebibliography}

\newpage
\appendix

\section{Gibbs Sampling Algorithm for B-DeepNoise}

\Cref{thm:distribution-v} provides
the full conditional posterior distribution
of the pre-activation latent noise variables.
In \Cref{thm:distribution-others},
we derive the counterparts for the other model parameters,
which are necessary for developing
a complete Gibbs sampling algorithm for B-DeepNoise.

\label{sec:gibbs}
\begin{proposition}
  \label{thm:distribution-others}
  Model parameters
  $\bfu^{(n)}_l$,
  $\bfbeta_l, \bfgamma_l$,
  $\bfrho^2_l, \bfxi^2_l$,
  $\bftau^2_l, \bfsigma^2_l$
  have the following full conditional posterior distributions:
  \begin{enumerate}
    \item Post-activation latent random noise:
      \begin{align}
        \label{eq:distribution-u}
        \bfu^{(n)}_{l}
        \Bigm| \rest 
        \sim
        \Normal(\bfmu_{\text{post},l,n}, \bfSigma_{\text{post},l,n}),
      \end{align}
      where
      \begin{align*}
        \bfmu_{\text{post}, l, n}
        = &
        (\bfbeta_l \bfbeta_l^\top
        + 
        \tilde{\bfSigma}^{-1}_{\text{post}, l, n})^{-1}
        [
        \tilde{\bfSigma}^{-1}_{\text{post}, l, n} h(\bfv_{l-1}^{(n)})
        +
        \bfbeta_l^\top (\bfv_l^{(n)} - \bfgamma_l)
        ]
        \\
        \bfSigma_{\text{post}, l, n}
        = &
        (\bfbeta_l^\top \diag[\bftau^{-2}_{l}] \bfbeta_l
        +
        \tilde{\bfSigma}^{-1}_{\text{post}, l, n})^{-1}
        ,\qquad
        \tilde{\bfSigma}_{\text{post}, l, n}
        =
        \diag[\bfsigma_{l-1}^2]
      \end{align*}
    \item Weight and bias parameters:
      \begin{align}
        \label{eq:distribution-betagamma}
        (\bfbeta_{l,k}, \gamma_{l,k})
        \Bigm| \rest
        \sim
        \Normal(
            \bfmu_{\text{WB}, l, k},
            \bfSigma_{\text{WB}, l, k})
      \end{align}
      where
      \begin{align*}
        \bfmu_{\text{WB}, l, k}
        = &
        (\bar{\bfu}_l \bar{\bfu}_l^\top
        + 
        \tilde{\bfSigma}^{-1}_{\text{WB}, l, k})^{-1}
        \bfv_{l,k} \bar{\bfu}_l^\top 
        \\
        \bfSigma_{\text{WB}, l, k}
        = &
        (\tau^{-2}_{l,k} \bar{\bfu}_l \bar{\bfu}_l^\top
        +
        \tilde{\bfSigma}^{-1}_{\text{WB}, l, k})^{-1},
        &
        \tilde{\bfSigma}_{\text{WB}, l, k}
        = &
        \diag(\bfrho_{l,k}^2, \xi_{l,k}^2)
        \\
        \bar{\bfu}_l
        = &
        (\bfu_l, \bm{1}),
        &
        \bfu_{l} = & \Big[ \bfu_{l}^{(n)} \Big]_{n=1}^N
      \end{align*}
    \item Variance parameters:
    \begin{align}
      \label{eq:distribution-tau}
      \tau^2_{l,k} \Bigm| \rest
      \sim &
      \InvGamma \left(
          a_{\text{postact}} + \frac{1}{2} N,\;
          b_{\text{postact}} + \frac{1}{2} \|
              \bfv_{l,k} - \bfbeta_{l,k}^\top \bfu_l - \gamma_{l,k}
          \|_2^2 \right)
      \\
      \label{eq:distribution-sigma}
      \sigma^2_{l,k} \Bigm| \rest
      \sim &
      \InvGamma \left(
          a_{\text{preact}} + \frac{1}{2} N,\;
          b_{\text{preact}} + \frac{1}{2} \|\bfu_{l+1,k} - h(\bfv_{l,k})\|_2^2
      \right)
      \\
      \label{eq:distribution-rho}
      \rho^2_{l,k,k'} \Bigm| \rest
      \sim &
      \InvGamma \left(
          a_{\text{weight}} + \frac{1}{2},\;
          b_{\text{weight}} + \frac{1}{2} \beta_{l,k,k'}^2
      \right)
      \\
      \label{eq:distribution-xi}
      \xi^2_{l,k} \Bigm| \rest
      \sim &
      \InvGamma \left(
          a_{\text{bias}} + \frac{1}{2},\;
          b_{\text{bias}} + \frac{1}{2} \gamma_{l,k}^2
      \right)
    \end{align}
    where
    \[
      \bfv_{l,k} = \Big[ v_{l,k}^{(n)} \Big]_{n=1}^N
      ,\qquad
      \bfu_{l,k} = \Big[ u_{l,k}^{(n)} \Big]_{n=1}^N
      ,\qquad
      \bfu_{l} = \Big[ \bfu_{l}^{(n)} \Big]_{n=1}^N
    \]
  \end{enumerate}
\end{proposition}

The Gibbs Sampling algorithm for B-DeepNoise is presented in \Cref{alg:sampling}.
\begin{algorithm}
  \caption{
    Posterior Sampler for Bayesian Deep Noise Neural Network
    (B-DeepNoise) \label{alg:sampling}
  }
  \SetAlgoLined
  \KwIn{
    training features and targets $[\bfx^{(n)}, \bfy^{(n)}]_{n=1}^{N_{\train}}$,
    testing features $[\bfx^{(n)}]_{n=1}^{N_{\test}}$,
    number of posterior samples $M$,
    number of realizations of the predictive distribution $R$.
  }
  \KwOut{
    posterior predictive samples for testing targets
    $[[[\hat{\bfy}^{(n)}_{m,r}]_{r=1}^R]_{m=1}^M]_{n=1}^{N_{\test}}$
  }
  Randomly initialize
    $
      [\bfbeta_{l}]_{l=0}^L, \;
      [\bfgamma_{l}]_{l=0}^L, \;
      [\bftau^2_{l}]_{l=0}^L, \;
      [\bfsigma^2_{l}]_{l=0}^L, \;
      [[\bfu^{(n)}_{l}]_{l=1}^L]_{n=1}^{N_{\train}}, \;
      [[\bfv^{(n)}_{l}]_{l=0}^{L-1}]_{n=1}^{N_{\train}}
    $ \;
  $[\bfu^{(n)}_0]_{n=1}^{N_{\train}} \leftarrow [\bfx^{(n)}]_{n=1}^{N_{\train}}$, \hspace{1em}
  $[\bfv^{(n)}_L]_{n=1}^{N_{\train}} \leftarrow [\bfy^{(n)}]_{n=1}^{N_{\train}}$ \;
  \For{$m \leftarrow 1, \ldots, M$}{
    \For{$l \leftarrow 0, \ldots, L$}{
      \For{$n \leftarrow 1, \ldots, N_{\train}$}{
        Sample $\bfv^{(n)}_l \mid \bfu^{(n)}_{l}, \bfu^{(n)}_{l+1}, \bfbeta_{l}, \bfgamma_{l}, \bfsigma^2_{l}, \bftau^2_{l}
$ by \Cref{eq:distribution-v} \;
        Sample $\bfu^{(n)}_{l+1} \mid \bfv^{(n)}_{l}, \bfv^{(n)}_{l+1}, \bfbeta_{l+1}, \bfgamma_{l+1}, \bfsigma^2_{l+1}, \bftau^2_{l+1}$ by \Cref{eq:distribution-u} \;
      }
      Sample $\bfbeta_l, \bfgamma_l \mid \bfu_{l}, \bfv_{l}, \bftau^2_{l}, \bfrho^2_{l}, \bfxi^2_{l}$ by \Cref{eq:distribution-betagamma}\;
      Sample $\bftau^2_l, \bfsigma^2_l, \bfrho^2_l, \bfxi^2_l \mid \bfv_{l}, \bfu_{l}, \bfbeta_{l}, \bfgamma_{l}, \bfu_{l+1}$
      by \Cref{eq:distribution-tau,eq:distribution-sigma,eq:distribution-rho,eq:distribution-xi}\;
    }
    $
      \hat{\bfbeta}_{m} \leftarrow [\bfbeta_{l}]_{l=0}^L, \;
      \hat{\bfgamma}_{m} \leftarrow [\bfgamma_{l}]_{l=0}^L, \;
      \hat{\bftau}^2_{m} \leftarrow [\bftau^2_{l}]_{l=0}^L, \;
      \hat{\bfsigma}^2_{m} \leftarrow [\bfsigma^2_{l}]_{l=0}^L
    $ \;
  }
  \For{$n \leftarrow 1, \ldots, N_{\test}$}{
    \For{$m \leftarrow 1, \ldots, M$}{
      \For{$r \leftarrow 1, \ldots, R$}{
        Sample $\bfy^{(n)}_{m} \mid \bfx^{(n)}, \hat{\bfbeta}_m, \hat{\bfgamma}_m, \hat{\bftau}^2_m, \hat{\bfsigma}^2_m$
        by \Cref{eq:u,eq:v} \;
        $\hat{\bfy}^{(n)}_{m,r} \leftarrow \bfy^{(n)}_{m}$
      }
    }
  }
\end{algorithm}

\end{document}